\documentclass{article}
\usepackage[margin=1in]{geometry}
\usepackage[utf8]{inputenc}
\usepackage[T1]{fontenc}
\usepackage{microtype}
\providecommand{\keywords}[1]{%
  \vspace{0.5em}\noindent\textbf{Keywords:} #1%
}
\usepackage{amsmath,amsfonts}
\usepackage{algorithmic}
\usepackage{algorithm}
\usepackage{array}
\usepackage{float}
\usepackage{placeins}
\usepackage{etoolbox}
\usepackage{booktabs}
\usepackage{dblfloatfix}
\usepackage{tabularx}
\usepackage{multirow}
\usepackage{makecell}

\usepackage{subfig} 
\usepackage{textcomp}
\usepackage{adjustbox}
\usepackage{amsmath}

\usepackage{url}
\usepackage{verbatim}
\usepackage{graphicx}
\usepackage{cite}
\pretocmd{\section}{\FloatBarrier}{}{}
\pretocmd{\subsection}{\FloatBarrier}{}{}
\pretocmd{\subsubsection}{\FloatBarrier}{}{}

\begin{document}

\title{PEARL: Path-Entity Aligned Relational Learning with Contextual Subgraphs for Inductive Knowledge Graph Completion}

\author{
Yunchi Yang \\
School of Mathematics, Shandong University \\
Jinan 250100, China \\
\texttt{ycyang@mail.sdu.edu.cn}
\and
Longlong Li\thanks{Corresponding author.} \\
School of Physical and Mathematical Sciences, Nanyang Technological University \\
Singapore 637371, Singapore \\
\texttt{longlong.li@ntu.edu.sg}
\and
Cunquan Qu\footnotemark[1] \\
Data Science Institute, Shandong University \\
Jinan 250100, China \\
\texttt{cqqu@sdu.edu.cn}
}

\maketitle

\begin{abstract}
Inductive knowledge graph completion (IKGC) aims to predict missing links involving entities unseen during training, requiring models to learn transferable relational and structural patterns. Existing subgraph- and path-based approaches often encode relational paths independently of their surrounding query subgraphs, although the predictive relevance of a path may vary across structural contexts. We propose PEARL, a Path–Entity Aligned Relational Learning framework that models paths as context-conditioned reasoning signals. PEARL constructs a query-specific contextual subgraph from the union of the query entities’ neighborhoods and uses a large language model (LLM)-guided retriever to distill semantically relevant paths. It then builds a bipartite interaction graph over paths, contextual entities, and a global subgraph representation, allowing path embeddings to adapt to local and global structural evidence. To suppress noise introduced by the enlarged context, PEARL employs a dual-view contrastive objective that promotes representation consistency under stochastic contextual perturbations. Experiments on WN18RR, FB15k-237, and NELL-995 show that PEARL obtains the best average Hits@10 among the compared IKGC methods on all three benchmarks. Ablation studies, efficiency analyses, and case studies further validate the contributions of contextual subgraph modeling, semantic path retrieval, path–entity interaction, and contrastive regularization.
\end{abstract}


\section{Introduction}

Knowledge Graphs (KGs) have become a fundamental infrastructure for numerous AI applications, including recommendation~\cite{intro_reco1,intro_reco2}, question answering~\cite{intro_ques1,intro_ques2}, and semantic search~\cite{intro_sem,intro_sem2}, by organizing factual knowledge as relational triples. Despite their large scale, real-world KGs are inherently incomplete, motivating extensive research on knowledge graph completion (KGC) to predict missing facts~\cite{intro_ikgc1,intro_ikgc2}. Most existing KGC models are developed under the transductive setting,
where all entities are assumed to be observed during training.
However, real-world KGs evolve continuously and frequently introduce new entities. For example, e-commerce knowledge graphs are constantly expanded with newly released products and brands. Retraining the entire model whenever new entities arrive is therefore impractical~\cite{intro_ecomm}. Consequently, inductive knowledge graph completion (IKGC)~\cite{adaprop,ingram,nbfnet} has attracted increasing attention, requiring models to learn transferable relational and structural patterns that generalize to unseen entities rather than relying on entity-specific embeddings.

Existing IKGC methods mainly exploit three types of reasoning evidence. The first is rule-based reasoning~\cite{drum,NeuralLP}, which learns transferable logical rules and naturally supports inductive inference through relation patterns. The second is subgraph evidence~\cite{s2dn,tact,compile,rmpi}, which captures the local structural context surrounding the query entities and enables reasoning through neighborhood interactions. The third is path evidence~\cite{snri,rpc,cats}, which provides explicit multi-hop relational chains connecting the query entities and offers complementary logical clues beyond local structural information. These three paradigms have significantly advanced inductive knowledge graph completion from different perspectives.

Effective inductive reasoning often benefits from exploiting a broad range of relational evidence surrounding the query entities, as additional triples may provide complementary structural and semantic clues for relation prediction. However, real-world knowledge graphs inevitably contain noisy or irrelevant information~\cite{kg_noise1,kg_noise2}, making not all available evidence equally useful for reasoning. Moreover, the predictive utility of a relational path may vary considerably across different structural contexts~\cite{LLM1}. For example, the same reasoning path $(\text{Company}_A \allowbreak ,\allowbreak \text{invests\_in} \allowbreak ,\allowbreak \text{Startup}_X \allowbreak ,\allowbreak \text{produces} \allowbreak ,\allowbreak \text{Product}_Y)$ may strongly support predicting an acquisition in one query, particularly when the surrounding structure reveals that $\text{Startup}_X$ heavily relies on $\text{Company}_A$'s supply chain and $\text{Company}_A$ holds a majority stake. In another query, however, the same path may provide much weaker evidence if $\text{Company}_A$ is merely a minor investor and $\text{Product}_Y$ is primarily sold to rival companies, suggesting a purely financial relationship rather than strategic alignment.

These characteristics give rise to two key challenges. First, how can richer relational and structural information be effectively incorporated while preserving the core semantics of the target triple and limiting the influence of noisy evidence? Second, local subgraphs often contain multiple relational paths connecting the query entities, some of which may be redundant or only weakly related to the target relation~\cite{LLM2}. Since the semantic importance of a path may further vary across different local structural environments, it remains challenging to identify informative reasoning paths and appropriately associate them with their surrounding context for query-specific reasoning.

To address these challenges, we propose PEARL, an IKGC framework based on Path-Entity Aligned Relational Learning. PEARL first constructs union-based contextual subgraphs to capture richer structural evidence. To alleviate the accompanying noise, a dual-view subgraph contrastive objective is introduced to preserve invariant structural semantics. Meanwhile, an LLM-guided path retrieval module ranks candidate paths and retains the most informative logical evidence. Finally, PEARL explicitly aligns paths with contextual entity nodes and the global subgraph representation through a Bipartite Graph Neural Network (BGNN), enabling context-conditioned path reasoning. Consequently, identical relational paths can be interpreted differently under different structural environments, leading to more robust inductive reasoning.

Our main contributions are summarized as follows:
\begin{itemize}
\item We propose PEARL, a unified framework for inductive knowledge graph completion that jointly models contextual subgraphs and relational paths to improve reasoning over unseen entities.

\item We develop an LLM-guided path retrieval and path--entity alignment strategy, together with contextual subgraph contrastive learning, to identify informative relational evidence and learn robust context-aware representations.

\item Extensive experiments on WN18RR, FB15k-237, and NELL-995
demonstrate that PEARL achieves strong overall performance and
the best average Hits@10 across all three benchmarks.
\end{itemize}

\section{Related Work}
Inductive knowledge graph completion aims to predict missing triples involving entities that are unseen during training. Unlike transductive link prediction, inductive reasoning requires models to learn transferable relational patterns rather than memorizing entity-specific embeddings~\cite{grail}. Existing methods can be broadly categorized into rule-based methods, subgraph-based methods, and subgraph \& path-based methods.

\subsection{Rule-based Methods}Rule-based methods are naturally suitable for inductive reasoning because they focus on learning logical patterns that are independent of specific entities. NeuralLP~\cite{NeuralLP} introduces an end-to-end differentiable framework that learns logical rules through differentiable reasoning operators. DRUM~\cite{drum} further improves expressiveness by introducing virtual relations and supporting rules of varying lengths through parameter sharing. Compared with traditional rule mining approaches such as AMIE~\cite{amle} and RuleN~\cite{relen}, differentiable rule learning methods provide better scalability and end-to-end optimization. Although rule-based methods offer strong interpretability and generalization to unseen entities, they often struggle to capture complex structural semantics and higher-order interactions within local graph neighborhoods.

\subsection{Subgraph-based Methods}Subgraph-based methods perform reasoning by extracting and encoding local graph structures surrounding the query entities. GraIL~\cite{grail} first introduced enclosing subgraph extraction for inductive link prediction, demonstrating the effectiveness of local structural patterns for reasoning over unseen entities. CoMPILE~\cite{compile} further enhances subgraph representations through relation-aware message passing, while TACT~\cite{tact} incorporates topological relation patterns to capture richer structural semantics.

To improve subgraph quality, LCILP~\cite{lcilp} employs a Personalized PageRank (PPR)-based clustering strategy to mitigate hub-node noise and extract more informative local neighborhoods. MINES~\cite{mines} introduces message intercommunication mechanisms over neighbor-enhanced subgraphs to strengthen information propagation. More recent methods such as RMPI~\cite{rmpi} and S$^2$DN~\cite{s2dn} further improve inductive reasoning through relational message passing and multi-view subgraph modeling, enabling more effective exploitation of structural semantics.

\subsection{Subgraph \& Path-based Methods}

Beyond local subgraph structures, explicit relational paths provide complementary multi-hop evidence by capturing the relation sequences that connect query entities~\cite{paths1,paths2}. NBFNet~\cite{nbfnet} formulates relational reasoning as a generalized Bellman--Ford process and learns to aggregate path evidence through neural message passing, achieving strong performance on both transductive and inductive link prediction tasks. SNRI~\cite{snri} incorporates neighboring relational information into subgraph representations to better capture the semantics of unseen entities. RPC-IR~\cite{rpc} further enhances relational path representations through contrastive learning and first-order rule induction, improving the quality of logical reasoning evidence. More recently, CATS~\cite{cats} introduces a context-aware framework that leverages large language models (LLMs) to assess query triples using latent type constraints, relational paths, and neighboring facts extracted from local subgraphs.



\begin{figure*}[t]
    \centering
    \includegraphics[width=\linewidth]{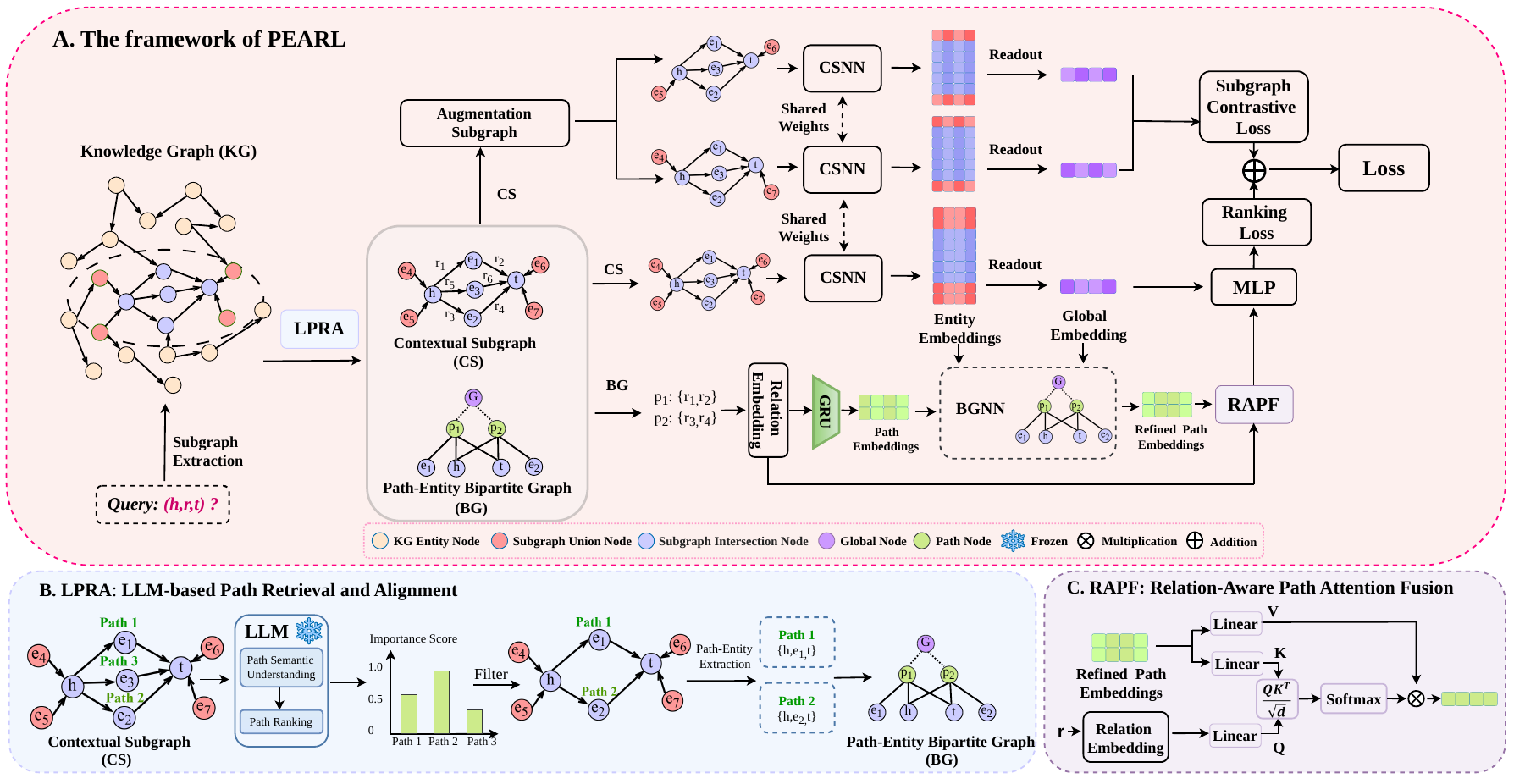}
    \caption{
Overall framework of PEARL. 
(A) Given a query triple, PEARL first extracts a contextual subgraph  from the knowledge graph. 
The proposed LLM-based Path Retrieval and Alignment (LPRA) module then retrieves informative relational paths and constructs the path--entity bipartite graph. 
The contextual subgraph is subsequently encoded by the Contextual Subgraph Neural Network (CSNN) to obtain contextual entity representations and a global subgraph representation. 
Subgraph Contrastive Learning (SCL) further regularizes the CSNN representations by enforcing consistency between two augmented contextual subgraph views. 
The resulting contextual representations are incorporated into the path--entity bipartite graph, where the Bipartite Graph Neural Network (BGNN) performs context-aware interactions between relational paths and structural information. 
Finally, the Relation-Aware Path Attention Fusion (RAPF) module aggregates the refined path representations for prediction. 
(B) Detailed illustration of the proposed LPRA module. 
(C) Detailed illustration of the proposed RAPF module.
}
    \label{fig:model}
\end{figure*}

\section{Preliminary}

\subsubsection{Knowledge graph}
A knowledge graph is a tuple $\mathcal{G}=(\mathcal{E},\mathcal{R},\mathcal{T})$, where $\mathcal{E}$ and $\mathcal{R}$ are finite sets of entities and relation types, respectively, and $\mathcal{T}\subseteq\mathcal{E}\times\mathcal{R}\times\mathcal{E}$ is the set of observed triples. Each triple $(h,r,t)\in\mathcal{T}$ represents a directed edge of type $r$ from the head entity $h$ to the tail entity $t$.

\subsubsection{Contextual Subgraph}
Subgraph-based IKGC~\cite{grail} reasons over a local graph associated with each query triple $q=(h,r_q,t)$. We instantiate this locality using undirected $k$-hop neighborhoods. Edge directions are ignored only when computing graph distances. Let $\operatorname{dist}_{\mathcal{G}}(u,v)$ denote the shortest-path distance between $u$ and $v$ in the undirected view of $\mathcal{G}$. The $k$-hop neighborhood of an entity $v$ is
\begin{equation}
\mathcal{N}^{k}(v)
=
\left\{u\in\mathcal{E}\,\middle|\,\operatorname{dist}_{\mathcal{G}}(u,v)\leq k\right\}.
\end{equation}

Prior work often employs the enclosing subgraph $\mathcal{G}_{enc} = (\mathcal{V}_{enc}, \mathcal{E}_{enc})$,
\begin{equation}
\mathcal{V}_{enc}=\mathcal{N}^{k}(h)\cap\mathcal{N}^{k}(t),
\end{equation}
\begin{equation}
\mathcal{E}_{enc}
=
\left\{(u,r,v)\in\mathcal{T}\,\middle|\,u,v\in\mathcal{V}_{enc}\right\}.
\end{equation}

We employ the contextual subgraph $\mathcal{G}_{cs} = (\mathcal{V}_{cs}, \mathcal{E}_{cs})$ obtained from the union of the query endpoints' neighborhoods:
\begin{equation}
\mathcal{V}_{cs}=\mathcal{N}^{k}(h)\cup\mathcal{N}^{k}(t),
\end{equation}
\begin{equation}
\mathcal{E}_{cs}
=
\left\{(u,r,v)\in\mathcal{T}\,\middle|\,u,v\in\mathcal{V}_{cs}\right\}.
\end{equation}
Thus $\mathcal{V}_{enc} \subseteq \mathcal{V}_{cs}$, and $\mathcal{G}_{enc}$ is the subgraph of $\mathcal{G}_{cs}$ induced by $\mathcal{V}_{enc}$.

\subsubsection{Link Prediction and Inductive Setting}
KGC ranks candidate entities for a query of the form $(h,r_q,?)$ or $(?,r_q,t)$. We learn a scoring function
$f:\mathcal{E}\times\mathcal{R}\times\mathcal{E}\to\mathbb{R}$ such that positive triples receive higher scores than corrupted triples. In the inductive setting considered here, the training and test entity sets are disjoint. The model must therefore rely on transferable structural and relational patterns rather than entity-specific identifiers~\cite{grail}.

\section{Methodology}

In this section, we present PEARL, as illustrated in Fig.~\ref{fig:model}, a unified framework for inductive link prediction that jointly models contextual structural evidence and multi-hop relational paths. PEARL constructs a richer query-specific context while preserving its core semantics. It then leverages the semantic reasoning capability
of a large language model to identify relational paths that are
semantically relevant to the target relation, thereby complementing purely structural signals.

PEARL consists of five main components. 
(i) The Contextual Subgraph Neural Network (CSNN) encodes the union-based contextual subgraph and produces both contextual entity representations and a global subgraph representation. 
(ii) The LLM-based Path Retrieval and Alignment (LPRA) module retrieves candidate paths from the contextual subgraph, ranks them according to their semantic relevance to the query relation, and constructs a path--entity bipartite graph by aligning the retained paths with their associated contextual entities and the global subgraph node. 
(iii) The Bipartite Graph Neural Network (BGNN) performs message passing over the path--entity bipartite graph to obtain refined path representations. 
(iv) The Relation-Aware Path Attention Fusion (RAPF) module adaptively aggregates the refined path representations according to the query relation. 
(v) The Subgraph Contrastive Learning (SCL) module promotes consistency between two augmented subgraphs to preserve query-relevant structural semantics and suppress noisy contextual perturbations.


\subsection{Contextual Subgraph Neural Network (CSNN)}

\subsubsection{Entity Node Representation Initialization}  
For inductive reasoning, each entity node $v\in\mathcal{V}_{cs}$ is represented by structural and relational features rather than a fixed entity identifier. Let $\mathbf{z}_v\in\mathbb{R}^{d_z}$ be the double-radius node label:
\[
\mathbf{z}_v
=
\left[
\operatorname{onehot}\!\left(\operatorname{dist}_{\mathcal{G}}(v,h)\right)
\parallel
\operatorname{onehot}\!\left(\operatorname{dist}_{\mathcal{G}}(v,t)\right)
\right],
\]
where $d_z$ is the dimension of the concatenated distance encoding. The initial entity node representation is
\begin{equation}
\mathbf{m}_v^{(0)}
=
\sigma\!\left(\mathbf{W}_0
\left[\mathbf{z}_v\parallel\mathbf{s}_v\right]+\mathbf{b}_0\right),
\qquad
\mathbf{W}_0\in\mathbb{R}^{d\times(d_z+d)}.
\end{equation}
Here, $\mathbf{s}_v\in\mathbb{R}^{d}$ is a query-aware aggregation of the relation embeddings incident to $v$:
\begin{equation}
\begin{aligned}
\mathbf{s}_v
&=
\sum_{r\in\mathcal{R}(v)}\alpha_{v,r}\mathbf{e}_r^{(0)},\\
\alpha_{v,r}
&=
\frac{\exp\!\left((\mathbf{e}_r^{(0)})^\top\mathbf{e}_{r_q}^{(0)}/\sqrt{d}\right)}
{\sum_{r'\in\mathcal{R}(v)}
\exp\!\left((\mathbf{e}_{r'}^{(0)})^\top\mathbf{e}_{r_q}^{(0)}/\sqrt{d}\right)}.
\end{aligned}
\end{equation}
Here, $\mathcal{R}(v)$ denotes the set of relation types incident to $v$, and $\mathbf{e}_{r_q}^{(0)}$ is the initial trainable embedding of the query relation $r_q$.

\subsubsection{Relation-aware Message Passing}  
To capture higher-order structural patterns and relational semantics, the node and relation representations are updated iteratively. At layer $\ell\in\{1,\ldots,L\}$, the update for node $i$ is
\begin{align}
\mathbf{c}_{ijr}^{(\ell)}
&=
\sigma_1\!\left(
\mathbf{W}_1^{(\ell)}
\left[
\mathbf{m}_i^{(\ell-1)}
\parallel\mathbf{m}_j^{(\ell-1)}
\parallel\mathbf{e}_r^{(\ell-1)}
\right]
+\mathbf{b}_1^{(\ell)}
\right),\\
\beta_{ijr}^{(\ell)}
&=
\sigma_2\!\left(
(\mathbf{w}_2^{(\ell)})^\top\mathbf{c}_{ijr}^{(\ell)}
+b_2^{(\ell)}
\right),\\
\mathbf{m}_i^{(\ell)}
&=
\sum_{r\in\mathcal{R}}
\sum_{j\in\mathcal{N}_r(i)}
\beta_{ijr}^{(\ell)}
\mathbf{W}_{r}^{(\ell)}
\phi\!\left(\mathbf{m}_j^{(\ell-1)},\mathbf{e}_r^{(\ell-1)}\right),\\
\mathbf{e}_r^{(\ell)}
&=
\mathbf{U}^{(\ell)}\mathbf{e}_r^{(\ell-1)}.
\end{align}
Here, $\mathcal{N}_r(i)=\{j\mid(j,r,i)\in\mathcal{E}_{cs}\text{ or }(i,r,j)\in\mathcal{E}_{cs}\}$ is the relation-specific neighborhood of $i$, $\phi(\cdot,\cdot)$ composes node and relation representations, and $\beta_{ijr}^{(\ell)}$ controls the contribution of the message from $j$ to $i$. The matrix $\mathbf{W}_{r}^{(\ell)}$ is relation specific, whereas $\mathbf{U}^{(\ell)}$ updates relation embeddings shared across relation types.

\subsubsection{Global Subgraph Embedding}
Let $\mathbf{M}^{(L)}=[\mathbf{m}_v^{(L)}]_{v\in\mathcal{V}_{cs}}
\in \mathbb{R}^{|\mathcal{V}_{cs}|\times d}$
denote the entity node embeddings obtained after $L$ layers of message passing, where
$\mathbf{m}_v^{(L)}\in\mathbb{R}^{d}$ is the embedding of node $v$.
Following CoMPILE~\cite{compile}, we further employ a Gated Recurrent Unit (GRU)~\cite{gru}
to refine the propagated node embeddings. The gating mechanism adaptively preserves
informative structural features while suppressing redundant information accumulated during
message passing. Specifically,
\begin{equation}
\widetilde{\mathbf{M}}^{(L)}
=
\operatorname{\textbf{GRU}}\!\left(\mathbf{M}^{(L)}\right)
=
[\widetilde{\mathbf{m}}_v^{(L)}]_{v\in\mathcal{V}_{cs}},
\end{equation}
where
$\widetilde{\mathbf{M}}^{(L)}
\in \mathbb{R}^{|\mathcal{V}_{cs}|\times d}$
denotes the final node embedding matrix and
$\widetilde{\mathbf{m}}_v^{(L)}\in\mathbb{R}^{d}$
is the final embedding of node $v$.
The global embedding of the contextual subgraph is then obtained using an average
readout function:
\begin{equation}
\mathbf{m}_{\mathcal{G}_{cs}}
=
\frac{1}{|\mathcal{V}_{cs}|}
\sum_{v\in\mathcal{V}_{cs}}
\widetilde{\mathbf{m}}_v^{(L)}
\in \mathbb{R}^{d}.
\end{equation}

\begin{table}[t]
\centering
\caption{Path Count Statistics on Various FB15k-237 Subsets.}
\label{tab:path_distribution}
\setlength{\tabcolsep}{8pt}
\renewcommand{\arraystretch}{1.1}
\begin{tabular}{cccc}
\toprule
\textbf{Version} & \textbf{Avg. Paths} & \textbf{$>$3 Paths (\%)} & \textbf{$>$10 Paths (\%)} \\
\midrule
V1 & 2.7  & 16.02 & 4.05 \\
V2 & 6.6  & 31.71 & 14.65 \\
V3 & 10.5 & 41.84 & 22.43 \\
V4 & 15.1 & 47.64 & 29.19 \\
\bottomrule
\end{tabular}
\end{table}

\subsection{LLM-based Path Retrieval and Alignment (LPRA)}

While the CSNN captures rich contextual structural information, simply extracting more paths from the contextual subgraph does not necessarily improve reasoning. Local subgraphs may contain multiple candidate paths with varying relevance to the query triple, and treating them equally can dilute informative evidence and introduce noisy reasoning signals.

To address this limitation, PEARL introduces an LLM-based Path Retrieval and Alignment (LPRA) module. LPRA leverages the semantic reasoning capability and relational knowledge of large language models to evaluate the semantic compatibility between each candidate path and the query triple.
This provides semantic guidance beyond purely structural evidence. The most informative paths are then retained as reliable relational evidence for subsequent path modeling.

We first perform a breadth-first search on the contextual subgraph $\mathcal{G}_{cs}$ to extract a candidate path set
$
\mathcal{P}_{\mathrm{cand}}
=
\{p_1,\ldots,p_N\}.
$
Each path $p_m$ is defined as a sequence of entities and relations connecting the head and tail entities:
\begin{equation}
p_m = (v_{m,0} = h, r_{m,1}, v_{m,1}, \ldots, r_{m,L_m}, v_{m,L_m} = t),
\end{equation}
where $L_m$ denotes the length of path $p_m$.

Although these candidate paths provide diverse multi-hop relational evidence, many of them are redundant or only weakly related to the target relation~\cite{LLM1,LLM2}. As shown in Table~\ref{tab:path_distribution}, the number of candidate paths grows substantially across different inductive settings of the FB15k-237 dataset~\cite{fb15}. While the average number of paths is only 2.7 in V1, it increases to 15.1 in V4, making it increasingly difficult to identify truly informative reasoning chains. Directly utilizing all candidate paths would not only introduce noisy reasoning signals but also increase the computational burden of downstream path modeling.

To improve the quality of path reasoning, we employ Qwen3-4B-Instruct~\cite{Qwen3} as an LLM-based path retriever~\cite{path_rank1,path_rank2}. The detailed prompt design is provided in supplemental material A. Given the query triple $q$ and the candidate path set $\mathcal{P}_{\mathrm{cand}}=\{p_1,\ldots,p_N\}$, the retriever evaluates the semantic support of each candidate path for the triple and assigns it an importance score:
\begin{equation}
s_{q,m}
=
\operatorname{\textbf{LLMScore}}
\left(
q,p_m
\right),
\qquad
p_m\in\mathcal{P}_{\mathrm{cand}},
\end{equation}
where $s_{q,m}$ denotes the semantic importance score of path $p_m$ for query $q$. We rank all candidate paths in descending order of their importance scores and denote the resulting index permutation as
$\pi_q=(\pi_q(1),\ldots,\pi_q(N))$.

According to the ranking results, we retain the top-$M$ candidate paths $\mathcal P_q
=
\{p_{\pi_q(1)},\ldots,p_{\pi_q(M)}\}.
$
Each retained path is explicitly aligned with the contextual entity nodes appearing along it, thereby establishing path--entity correspondences.

Based on these paths, we construct a path--entity bipartite graph
$
\mathcal{B}_q=(\mathcal{U}_q,\mathcal{E}_{\mathcal{B}}).
$
Its node set is defined as
$
\mathcal{U}_q
=
\mathcal{V}_{cs}\cup\{g\}
\cup
\mathcal{P}_q,
$
where $\mathcal{V}_{cs}$ denotes the contextual entity nodes, $\mathcal{P}_q$ is the retained path set, and $g$ is a virtual node representing the global contextual subgraph.

The edge set is defined as
\[
\mathcal{E}_{\mathcal{B}}
=
\left\{
\{v,p_m\}
\,\middle|\,
v\in\mathcal V_{cs},\,
v\in p_m
\right\}
\cup
\left\{
\{g,p_m\}
\,\middle|\,
p_m\in\mathcal P_q
\right\},
\]
which contains two types of connections, namely path--entity (PE) edges and path--global (PG) edges. The path--entity edges connect each relational path with the contextual entities appearing on that path, while the path--global edges link every path to the global contextual node. 

\subsection{Bipartite Graph Neural Network (BGNN)}

BGNN performs message passing over the path--entity bipartite graph to model interactions among relational paths, contextual entity nodes, and the global contextual node. Through iterative information propagation, each path representation is adaptively refined according to its surrounding structural context.

\subsubsection{Path Embedding Initialization}

For each retained path
\[
p_m
=
(v_{m,0}{=}h,r_{m,1},v_{m,1},
\ldots,
r_{m,L_m},v_{m,L_m}{=}t)
\in\mathcal{P}_q,
\]
we encode its relation sequence using a GRU encoder to capture sequential dependencies and directional semantics:
\begin{equation}
\mathbf{p}_m^{(0)}
=
\textbf{GRU}\!\left(
\mathbf{e}_{r_{m,1}}^{(L)},
\ldots,
\mathbf{e}_{r_{m,L_m}}^{(L)}
\right)
\in\mathbb{R}^{d}.
\end{equation}

\subsubsection{Node Initialization}

The initial embeddings of the nodes in
$\mathcal{B}_q$ are defined as
\begin{equation}
\mathbf{z}_i^{(0)}
=
\begin{cases}
\widetilde{\mathbf{m}}_v^{(L)},
& i=v\in\mathcal{V}_{cs},\\[2pt]
\mathbf{m}_{\mathcal{G}_{cs}},
& i=g,\\[2pt]
\mathbf{p}_m^{(0)},
& i=p_m\in\mathcal{P}_q,
\end{cases}
\end{equation}
where $\widetilde{\mathbf{m}}_v^{(L)}$ denotes the contextual entity
embedding produced by CSNN,
$\mathbf{m}_{\mathcal{G}_{cs}}$ is the global contextual subgraph
embedding, and $\mathbf{p}_m^{(0)}$ is the initial embedding of
path $p_m$.

\subsubsection{Message Passing with R-GAT}

To preserve the distinct semantic roles of different interactions during information propagation, BGNN employs a multi-head Relational Graph Attention Network (R-GAT)~\cite{gat}.

Let
$
\mathcal{T}_{\mathcal B}
=
\{\mathrm{PE},\mathrm{PG}\}
$
denote the interaction-type set of the bipartite graph.
Accordingly, the type-specific neighborhood of node $i$ is defined as
\begin{equation}
\mathcal{N}_{\tau}^{\mathcal{B}}(i)
=
\left\{
j\in\mathcal U_q
\ \middle|\
\{i,j\}\in\mathcal E_{\mathcal B},
\ \tau(i,j)=\tau
\right\}.
\end{equation}

The interaction type of each edge is determined by the node categories:
\begin{equation}
\tau(i,j)=
\begin{cases}
\mathrm{PE},
&
(i\in\mathcal{V}_{cs}, j\in\mathcal{P}_q)
\ \text{or}\
(i\in\mathcal{P}_q, j\in\mathcal{V}_{cs}),
\\
\mathrm{PG},
&
(i=g,j\in\mathcal{P}_q)
\ \text{or}\
(i\in\mathcal{P}_q,j=g).
\end{cases}
\end{equation}

For each edge $(i,j)$ with interaction type $\tau$, we first compute the
attention score to measure the importance of node $j$ to node $i$ under the
specific interaction type:
\begin{align}
e_{ij,\tau}^{(\ell,a)}
=
\operatorname{LeakyReLU}
\Big(
(\mathbf a_\tau^{(\ell,a)})^\top
[
\mathbf h_{i,\tau}^{(\ell,a)}
\parallel
\mathbf h_{j,\tau}^{(\ell,a)}
]
\Big),
\end{align}
where
$
\mathbf h_{i,\tau}^{(\ell,a)}
=
\mathbf W_\tau^{(\ell,a)}
\mathbf z_i^{(\ell-1)}
$
denotes the type-specific transformed representation, and
$\mathbf a_\tau^{(\ell,a)}$ is a learnable attention vector.

The attention scores are normalized over all neighbors of node $i$ to obtain
the attention coefficients:
\begin{align}
\alpha_{ij,\tau}^{(\ell,a)}
=
\frac{
\exp(e_{ij,\tau}^{(\ell,a)})
}{
\displaystyle
\sum_{\tau'\in\mathcal T_{\mathcal B}}
\sum_{j'\in\mathcal{N}_{\tau'}^{\mathcal{B}}(i)}
\exp(e_{ij',\tau'}^{(\ell,a)})
}.
\end{align}

Using the obtained attention coefficients, the embedding of node $i$ is
updated through multi-type message aggregation:
\begin{align}
\mathbf z_i^{(\ell)}
=
\mathop{\Big\Vert}_{a=1}^{H}
\sigma\!\left(
\sum_{\tau\in\mathcal T_{\mathcal B}}
\sum_{j\in\mathcal{N}_{\tau}^{\mathcal{B}}(i)}
\alpha_{ij,\tau}^{(\ell,a)}
\mathbf W_\tau^{(\ell,a)}
\mathbf z_j^{(\ell-1)}
\right),
\end{align}
where $H$ denotes the number of attention heads, and
$\mathbf W_\tau^{(\ell,a)}
\in \mathbb{R}^{\frac{d}{H}\times d}$
is the transformation matrix associated with interaction type $\tau$.
After $L_B$ R-GAT layers, the refined embedding of each
retained relational path is obtained as
\begin{equation}
\widetilde{\mathbf{p}}_m
=
\mathbf z_{p_m}^{(L_B)}.
\end{equation}

Through type-aware attention propagation over the path--entity bipartite
graph, each path embedding selectively aggregates local contextual
entity information and global subgraph semantics, enabling identical
relational paths to be interpreted differently under different structural
environments.

\subsection{Relation-Aware Path Attention Fusion (RAPF)}

The refined path embeddings $\{ \widetilde{\mathbf{p}}_m \}_{m=1}^M$ produced by BGNN encode diverse multi-hop relational semantics between the query entities. To identify the most informative logical evidence for the query relation $r_q$, we employ a Scaled Dot-Product Attention mechanism to aggregate these paths into a relation-aware path representation $\mathbf{p}_{\mathcal{G}_{cs}} \in \mathbb{R}^d$.

Specifically, the final query-relation embedding $\mathbf{e}_{r_q}^{(L)}\in\mathbb{R}^d$ is used as the query, while the refined path embeddings serve as keys and values. The projected query, key, and value vectors are
\begin{equation}
\small  
\mathbf{q}_{r_q}=\mathbf{W}^{Q}\mathbf{e}_{r_q}^{(L)},\qquad
\mathbf{k}_m=\mathbf{W}^{K}\widetilde{\mathbf{p}}_m,\qquad
\mathbf{v}_m=\mathbf{W}^{V}\widetilde{\mathbf{p}}_m,
\end{equation}
where $\mathbf{q}_{r_q},\mathbf{k}_m, \mathbf{v}_m\in\mathbb{R}^{d}$. The matrices $\mathbf{W}^{Q}$, $\mathbf{W}^{K}$, and $\mathbf{W}^{V}$ are learnable projections.

The attention weight assigned to each path is computed as follows:
\begin{equation}
\alpha_m
=
\frac{
\exp \left(
\frac{\mathbf{q}_{r_q}^{\top}\mathbf{k}_{m}}{\sqrt{d}}
\right)
}
{
\sum_{i=1}^{M}
\exp \left(
\frac{\mathbf{q}_{r_q}^{\top}\mathbf{k}_{i}}{\sqrt{d}}
\right)
}.
\end{equation}

Finally, the aggregated path embedding is obtained as:
\begin{equation}
\mathbf{p}_{\mathcal{G}_{cs}}
=
\sum_{m=1}^{M}
\alpha_m\mathbf{v}_m.
\end{equation}

﻿
﻿
﻿
\subsection{Subgraph Contrastive Learning (SCL)}

Existing inductive reasoning methods typically construct enclosing subgraphs based on the intersection of the $k$-hop entity neighborhoods of the query entities, which preserves reliable shared structural information but may overlook complementary contextual semantics. In contrast, the proposed contextual subgraph $\mathcal{G}_{cs}$ incorporates nodes from the union of the two entity neighborhoods, enabling richer structural representation and broader contextual coverage. However, directly introducing all contextual entity nodes may also bring irrelevant or noisy structural signals, potentially weakening the core semantics associated with the query triple.

To balance structural enrichment and semantic consistency, PEARL adopts a dual-view subgraph augmentation contrastive learning strategy~\cite{GraphCL}. Specifically, we partition the entity node set $\mathcal{V}_{cs}$ into core nodes and contextual nodes:
\begin{equation}
\mathcal{V}_{enc}
=
\mathcal{N}^{k}(h)\cap\mathcal{N}^{k}(t),
\qquad
\mathcal{V}_{\Delta}
=
\mathcal{V}_{cs}\setminus\mathcal{V}_{enc},
\end{equation}
$\mathcal{V}_{enc}$ preserves the shared structural semantics of the query entities, while $\mathcal{V}_{\Delta}$ contains complementary contextual information.

Two stochastic subgraph views are then constructed by independently sampling from $\mathcal{V}_{\Delta}$ without replacement. Specifically, for each view $a \in \{1,2\}$, we randomly draw exactly $\lfloor |\mathcal{V}_{\Delta}| / 2 \rfloor$ nodes from $\mathcal{V}_{\Delta}$ to form $\mathcal{V}_{\Delta}^{(a)}$, with each subset being sampled uniformly among all possible subsets of that size. The entity node sets of the two subgraph views are then defined as
\begin{equation}
\mathcal{V}_{cs}^{(a)} = \mathcal{V}_{enc} \cup \mathcal{V}_{\Delta}^{(a)}, \qquad a \in \{1,2\},
\end{equation}
and each view $\mathcal{G}_{cs}^{(a)}$ is the subgraph of $\mathcal{G}_{cs}$ induced by $\mathcal{V}_{cs}^{(a)}$.

Since the two augmented subgraphs share the same core structure but differ in the sampled contextual nodes, the model is encouraged to learn embeddings that are invariant to contextual perturbations. For a mini-batch of $B$ query triples, the two augmented subgraphs are encoded by the same CSNN with shared parameters, producing
$\mathbf{m}_{cs,1}^{(b)}$ and $\mathbf{m}_{cs,2}^{(b)}$,
for the $b$-th query. We optimize the subgraph contrastive loss using the following one-directional InfoNCE objective~\cite{infonce}:
\begin{equation}
\small  
\mathcal{L}_{\mathrm{SCL}}
=
-\frac{1}{B}
\sum_{b=1}^{B}
\log
\frac{
\exp\!\left(
\operatorname{cos}
\left(
\mathbf{m}_{cs,1}^{(b)},
\mathbf{m}_{cs,2}^{(b)}
\right)
/\tau
\right)
}{
\sum_{c=1}^{B}
\exp\!\left(
\operatorname{cos}
\left(
\mathbf{m}_{cs,1}^{(b)},
\mathbf{m}_{cs,2}^{(c)}
\right)
/\tau
\right)
},
\end{equation}
where $\operatorname{cos}(\cdot,\cdot)$ denotes cosine similarity and $\tau>0$ is the temperature parameter. For the anchor embedding $\mathbf{m}_{cs,1}^{(b)}$, its counterpart $\mathbf{m}_{cs,2}^{(b)}$ forms the positive sample, while $\{\mathbf{m}_{cs,2}^{(c)}\}_{c\neq b}$ are treated as in-batch negative samples.

By maximizing consistency across two augmented subgraphs, the proposed contrastive objective encourages PEARL to focus on stable and query-relevant structural semantics while suppressing the influence of unstable contextual noise.

\subsection{Training and Optimization}

\subsubsection{Joint Scoring Function}  
To predict the plausibility of a query triple $q=(h,r_q,t)$, we first integrate the contextual subgraph embedding and the aggregated path embedding into a unified structural-semantic vector:
\begin{equation}
\widetilde{\mathbf{s}_q}
=
[
\mathbf{m}_{\mathcal{G}_{cs}}
\parallel
\mathbf{p}_{\mathcal{G}_{cs}}
].
\end{equation}

The final prediction score is computed based on the query entities, the query relation, and the fused structural-path representation:
\begin{equation}
f(h,r_q,t)
=
\mathbf{w}_s^\top
\left[
\widetilde{\mathbf{m}}_h^{(L)}
\parallel
\widetilde{\mathbf{m}}_t^{(L)}
\parallel
\mathbf{e}_{r_q}^{(L)}
\parallel
\widetilde{\mathbf{s}_q}
\right]
+b_s,
\end{equation}
where $\parallel$ denotes vector concatenation, $\widetilde{\mathbf{m}}_h^{(L)}$ and $\widetilde{\mathbf{m}}_t^{(L)}$ are the embeddings from CSNN, and $\mathbf{w}_s$ and $b_s$ are learnable scoring parameters.

\subsubsection{Objective Function}  
To train PEARL, for each positive triple
$q_n^+=(h_n,r_n,t_n)\in\mathcal{T}_{\mathrm{train}}$,
we construct a negative triple by corrupting either the head or the tail:
\begin{equation}
q_n^-\in
\left\{
(h_n',r_n,t_n),\,
(h_n,r_n,t_n')
\right\}.
\end{equation}
The prediction objective is optimized using a margin-based ranking loss:
\begin{equation}
\mathcal{L}_{\mathrm{task}}
=
\frac{1}{|\mathcal{T}_{\mathrm{train}}|}
\sum_{n=1}^{|\mathcal{T}_{\mathrm{train}}|}
\left[
\gamma
+f(q_n^-)
-f(q_n^+)
\right]_+,
\end{equation}
where $[x]_+=\max(0,x)$ and $\gamma>0$ denotes the margin.

To further improve the robustness and semantic consistency of contextual subgraph representations, we additionally incorporate the proposed subgraph contrastive loss $\mathcal{L}_{\text{SCL}}$. The overall training objective is defined as:
\begin{equation}
\mathcal{L}
=
\lambda_1\mathcal{L}_{\text{task}}
+\lambda_2\mathcal{L}_{\text{SCL}},
\end{equation}
where $\lambda_1,\lambda_2\geq0$ control the relative contributions of the two objectives.
\begin{table*}[t]
\centering
\fontsize{5.8pt}{6.3pt}\selectfont
\setlength{\tabcolsep}{2.4pt}
\renewcommand{\arraystretch}{1.5}  

\newcommand{\mainmean}[2]{#1\,$\pm$\,{\fontsize{4.8pt}{5pt}\selectfont #2}}
\newcommand{\mainbest}[2]{\textbf{#1\,$\pm$\,{\fontsize{4.8pt}{5pt}\selectfont #2}}}

\newcommand{\secmean}[2]{\underline{\mainmean{#1}{#2}}}
\newcommand{\secbest}[2]{\underline{\mainbest{#1}{#2}}}

\caption{Performance comparison on WN18RR, FB15k-237, and NELL-995 under the V1 and V3 inductive settings (\%). results are reported as mean ± standard deviation. Best results are in bold; second-best are underlined.}
\label{tab:main_results}
\resizebox{\textwidth}{!}{%
\begin{tabular}{l|cccc|cccc|cccc}
\toprule
\multirow{3}{*}{\textbf{Methods}}
& \multicolumn{4}{c|}{\textbf{WN18RR}}
& \multicolumn{4}{c|}{\textbf{FB15k-237}}
& \multicolumn{4}{c}{\textbf{NELL-995}} \\
\cmidrule(lr){2-5}\cmidrule(lr){6-9}\cmidrule(lr){10-13}
& \multicolumn{2}{c}{\textbf{V1}} & \multicolumn{2}{c|}{\textbf{V3}}
& \multicolumn{2}{c}{\textbf{V1}} & \multicolumn{2}{c|}{\textbf{V3}}
& \multicolumn{2}{c}{\textbf{V1}} & \multicolumn{2}{c}{\textbf{V3}} \\
\cmidrule(lr){2-3}\cmidrule(lr){4-5}
\cmidrule(lr){6-7}\cmidrule(lr){8-9}
\cmidrule(lr){10-11}\cmidrule(lr){12-13}
& Hits@10 & MRR & Hits@10 & MRR
& Hits@10 & MRR & Hits@10 & MRR
& Hits@10 & MRR & Hits@10 & MRR \\
\midrule
\multicolumn{13}{l}{\textbf{Rule-based Methods:}} \\
DRUM~\cite{drum}
& \mainmean{75.36}{0.0013} & \mainmean{48.67}{0.0006} & \mainmean{47.19}{0.0007} & \mainmean{30.73}{0.0014}
& \mainmean{52.05}{0.0025} & \mainmean{33.29}{0.0022} & \mainmean{48.56}{0.0020} & \mainmean{31.63}{0.0014}
& \mainmean{40.80}{0.0024} & \mainmean{40.76}{0.0003} & \mainmean{50.53}{0.0018} & \mainmean{31.41}{0.0017} \\
NeuralLP~\cite{NeuralLP}
& \mainmean{75.76}{0.0000} & \mainmean{45.83}{0.0019} & \mainmean{46.90}{0.0011} & \mainmean{30.28}{0.0007}
& \mainmean{49.56}{0.0024} & \mainmean{25.31}{0.0028} & \mainmean{48.05}{0.0017} & \mainmean{29.60}{0.0014}
& \mainmean{40.50}{0.0000} & \mainmean{40.60}{0.0001} & \mainmean{49.94}{0.0040} & \mainmean{29.45}{0.0024} \\
\midrule
\multicolumn{13}{l}{\textbf{Subgraph-based Methods:}} \\
GraIL~\cite{grail}
& \mainmean{82.04}{0.0035} & \mainmean{74.69}{0.0116} & \mainmean{59.72}{0.0027} & \mainmean{54.24}{0.0042}
& \mainmean{62.92}{0.0086} & \mainmean{45.69}{0.0053} & \mainmean{80.66}{0.0025} & \mainmean{59.62}{0.0028}
& \mainmean{54.80}{0.0057} & \mainmean{43.86}{0.0113} & \mainmean{87.00}{0.0051} & \mainmean{74.24}{0.0057} \\
CoMPILE~\cite{compile}
& \mainmean{83.24}{0.0033} & \mainmean{76.95}{0.0118} & \mainmean{59.75}{0.0037} & \mainmean{54.44}{0.0036}
& \mainmean{64.73}{0.0085} & \mainmean{48.01}{0.0050} & \secmean{83.91}{0.0036} & \secmean{63.81}{0.0041}
& \mainmean{55.50}{0.0054} & \mainmean{48.85}{0.0056} & \secmean{94.43}{0.0043} & \mainbest{79.44}{0.0053} \\
TACT~\cite{tact}
& \mainmean{82.45}{0.0041} & \mainmean{75.45}{0.0115} & \mainmean{58.60}{0.0005} & \mainmean{54.29}{0.0026}
& \mainmean{63.17}{0.0045} & \mainmean{48.55}{0.0065} & \mainmean{82.29}{0.0031} & \mainmean{62.81}{0.0040}
& \mainmean{53.50}{0.0040} & \mainmean{45.27}{0.0020} & \mainmean{93.82}{0.0028} & \mainmean{69.74}{0.0043} \\
RMPI~\cite{rmpi}
& \mainmean{82.45}{0.0031} & \mainmean{65.66}{0.0111} & \mainmean{58.84}{0.0058} & \mainmean{53.84}{0.0052}
& \secmean{66.10}{0.0029} & \mainmean{47.55}{0.0057} & \mainmean{83.06}{0.0040} & \mainmean{63.37}{0.0029}
& \mainmean{59.00}{0.0080} & \mainmean{48.87}{0.0059} & \mainmean{92.89}{0.0008} & \secmean{78.89}{0.0042} \\
LCILP~\cite{lcilp}
& \mainmean{88.29}{0.0024} & \mainmean{77.69}{0.0052} & \mainmean{67.27}{0.0023} & \secmean{61.11}{0.0039}
& \mainmean{59.26}{0.0021} & \mainmean{45.84}{0.0055} & \mainmean{82.77}{0.0014} & \mainmean{56.65}{0.0013}
& \mainmean{52.00}{0.0090} & \mainmean{46.57}{0.0024} & \mainmean{87.20}{0.0033} & \mainmean{69.38}{0.0060} \\
S$^2$DN~\cite{s2dn}
& \mainmean{86.43}{0.0089} & \secmean{78.62}{0.0119} & \secmean{69.01}{0.0089} & \mainmean{54.82}{0.0050}
& \mainmean{64.39}{0.0118} & \secmean{48.59}{0.0099} & \mainmean{82.94}{0.0033} & \mainmean{59.21}{0.0052}
& \mainmean{61.00}{0.0082} & \mainmean{49.69}{0.0135} & \mainmean{91.71}{0.0040} & \mainmean{71.83}{0.0034} \\
MINES~\cite{mines}
& \secmean{88.45}{0.0085} & \mainmean{75.51}{0.0095} & \mainmean{62.46}{0.0043} & \mainmean{55.44}{0.0030}
& \mainmean{61.51}{0.0065} & \mainmean{46.60}{0.0086} & \mainmean{80.31}{0.0072} & \mainmean{58.51}{0.0077}
& \mainmean{59.40}{0.0027} & \mainmean{49.77}{0.0076} & \mainmean{92.37}{0.0036} & \mainmean{64.20}{0.0040} \\
\midrule
\multicolumn{13}{l}{\textbf{Subgraph \& Path-based Methods:}} \\
NBFNet~\cite{nbfnet}
& \mainmean{83.24}{0.0034} & \mainmean{69.50}{0.0032} & \mainmean{54.46}{0.0132} & \mainmean{43.40}{0.0084}
& \mainmean{50.63}{0.0158} & \mainmean{30.49}{0.0100} & \mainmean{61.44}{0.0016} & \mainmean{35.27}{0.0037}
& \mainmean{57.70}{0.0075} & \mainmean{50.25}{0.0163} & \mainmean{49.15}{0.0040} & \mainmean{28.28}{0.0186} \\
CATS~\cite{cats}
& \mainmean{78.45}{0.0054} & \mainmean{19.38}{0.0115} & \mainmean{56.77}{0.0059} & \mainmean{35.24}{0.0053}
& \mainmean{54.14}{0.0109} & \mainmean{32.96}{0.0074} & \mainmean{59.53}{0.0053} & \mainmean{34.03}{0.0040}
& \mainbest{66.10}{0.0065} & \secmean{50.27}{0.0031} & \mainmean{94.19}{0.0052} & \mainmean{73.43}{0.0039} \\
SNRI~\cite{snri}
& \mainmean{85.84}{0.0060} & \mainmean{75.41}{0.0043} & \mainmean{57.68}{0.0027} & \mainmean{46.92}{0.0028}
& \mainmean{63.90}{0.0124} & \mainmean{44.76}{0.0082} & \mainmean{82.65}{0.0034} & \mainmean{59.99}{0.0045}
& \mainmean{55.50}{0.0108} & \mainmean{42.72}{0.0045} & \mainmean{93.56}{0.0028} & \mainmean{72.23}{0.0049} \\
RPC-IR~\cite{rpc}
& \mainmean{78.73}{0.0069} & \mainmean{69.18}{0.0060} & \mainmean{64.64}{0.0048} & \mainmean{55.08}{0.0044}
& \mainmean{62.53}{0.0085} & \mainmean{41.67}{0.0050} & \mainmean{80.80}{0.0067} & \mainmean{56.60}{0.0036}
& \mainmean{60.80}{0.0120} & \mainmean{46.33}{0.0142} & \mainmean{84.61}{0.0075} & \mainmean{71.94}{0.0019} \\
\midrule
PEARL
& \mainbest{93.48}{0.0048} & \mainbest{80.36}{0.0029} & \mainbest{81.25}{0.0022} & \mainbest{66.82}{0.0026}
& \mainbest{72.63}{0.0060} & \mainbest{50.55}{0.0115} & \mainbest{86.31}{0.0042} & \mainbest{66.37}{0.0030}
& \secmean{63.00}{0.0073} & \mainbest{50.77}{0.0188} & \mainbest{96.89}{0.0095} & \mainmean{75.83}{0.0077} \\
\bottomrule
\end{tabular}%
}
\end{table*}



\section{Experiments}

\subsection{Datasets and Evaluation}

We evaluate PEARL on three widely used IKGC datasets: WN18RR~\cite{wn18}, FB15k-237~\cite{fb15}, and NELL-995~\cite{nell}. Following GraIL~\cite{grail}, each dataset is divided into four inductive subsets (V1--V4) with increasing scales, where the
training and test graphs contain disjoint entity sets and are
constructed under a shared global relation vocabulary.

For evaluation, we adopt the standard filtered ranking protocol and report Hits@1, Hits@10, and Mean Reciprocal Rank (MRR), where higher values indicate better performance. All results are averaged over five independent runs. Detailed dataset statistics and evaluation settings are provided in supplemental material B.

\subsection{Baselines} 

To evaluate the effectiveness of PEARL, we compare it with representative IKGC methods from three categories:

\begin{itemize}
    \item \textbf{Rule-based Methods}: NeuralLP~\cite{NeuralLP}, DRUM~\cite{drum}.
    
    \item \textbf{Subgraph-based Methods}: GraIL~\cite{grail}, CoMPILE~\cite{compile}, TACT~\cite{tact}, RMPI~\cite{rmpi}, LCILP~\cite{lcilp}, S$^2$DN~\cite{s2dn}, and MINES~\cite{mines}.
    
    \item \textbf{Subgraph \& Path-based Methods}: NBFNet~\cite{nbfnet}, SNRI~\cite{snri}, RPC-IR~\cite{rpc}, and CATS~\cite{cats}.
\end{itemize}

Detailed descriptions of all baseline models are provided in supplemental material C.

\subsection{Implementation Details}
The number of message-passing layers is set to $L=3$ for CSNN and $L_B=3$ for BGNN. The maximum relational path length is set to $L_m=2$, and the R-GAT employs $H=2$ attention heads. The hidden dimension $d$ is fixed to 32 for all node, relation, and path embeddings. All experiments are implemented in PyTorch and conducted on a single NVIDIA RTX 4090 GPU. The remaining hyperparameters, including the number of retained paths $M$, learning rate, batch size, loss coefficients, and the subgraph size $k$, are selected according to validation performance. Detailed configurations and search ranges are provided in supplemental material E. The source code and implementation details are publicly available at \url{https://github.com/yychi-code/PEARL}.

\subsection{Results and Analysis}

The experimental results on WN18RR, FB15k-237, and NELL-995 are summarized in Table~\ref{tab:main_results}. Overall, PEARL achieves the strongest average performance across the evaluated settings. Compared with the strongest competing methods, PEARL achieves relative improvements of 8.67\%, 6.01\%, and
3.20\% in average Hits@10 over the strongest competing methods
on WN18RR, FB15k-237, and NELL-995, respectively, demonstrating its strong generalization ability for inductive knowledge graph completion.

Several observations can be drawn from the experimental results. Rule-based methods, including NeuralLP and DRUM, generally achieve relatively lower performance, suggesting that logical rule induction alone may not fully capture the rich structural dependencies surrounding unseen entities. 
Subgraph-based approaches consistently achieve stronger performance by explicitly exploiting neighborhood structures. GraIL establishes an effective inductive reasoning framework based on
enclosing subgraphs. Subsequent methods further enhance subgraph
representations from different perspectives. Specifically, CoMPILE
improves message passing, TACT models topological relation patterns,
SNRI incorporates semantic neighboring relations, RMPI develops
relational message passing, LCILP adopts PPR-based local clustering,
MINES introduces message intercommunication, and S$^2$DN employs
dual-view structural learning. NBFNet formulates multi-hop reasoning as differentiable path aggregation and achieves competitive performance across several datasets. More recently, CATS incorporates LLMs to leverage latent type constraints, neighboring facts, and relational paths, demonstrating the potential of LLM-assisted inductive reasoning. Nevertheless, these methods either primarily focus on structural representations or utilize relational paths without explicitly modeling their interactions with the surrounding contextual subgraph. Although RPC-IR further improves path representations through contrastive learning, it primarily focuses on relational paths and does not explicitly model their interactions with the surrounding structural context.

In contrast, PEARL jointly models contextual subgraphs and relational paths within a unified framework. The LLM-guided path retrieval module identifies informative reasoning paths, while BGNN enables context-aware interactions between paths and structural embeddings. Combined with contextual subgraph modeling and subgraph contrastive learning, PEARL effectively exploits both structural and logical evidence, resulting in strong and generally consistent performance across the benchmark datasets.

\subsection{Efficiency Analysis}

We further evaluate the computational efficiency of PEARL from both the training and inference perspectives. We select several competitive methods with strong performance on the FB15k-237 V1 and V3 datasets for efficiency comparison. To ensure a fair comparison, all models are evaluated under the same hardware environment and evaluation protocol. Training is performed on a single NVIDIA RTX 4090 GPU. For inference evaluation, preprocessing and downstream model inference are measured separately, with the latter conducted on a CPU.

\begin{figure}[t]
    \centering
    \includegraphics[width=\linewidth]{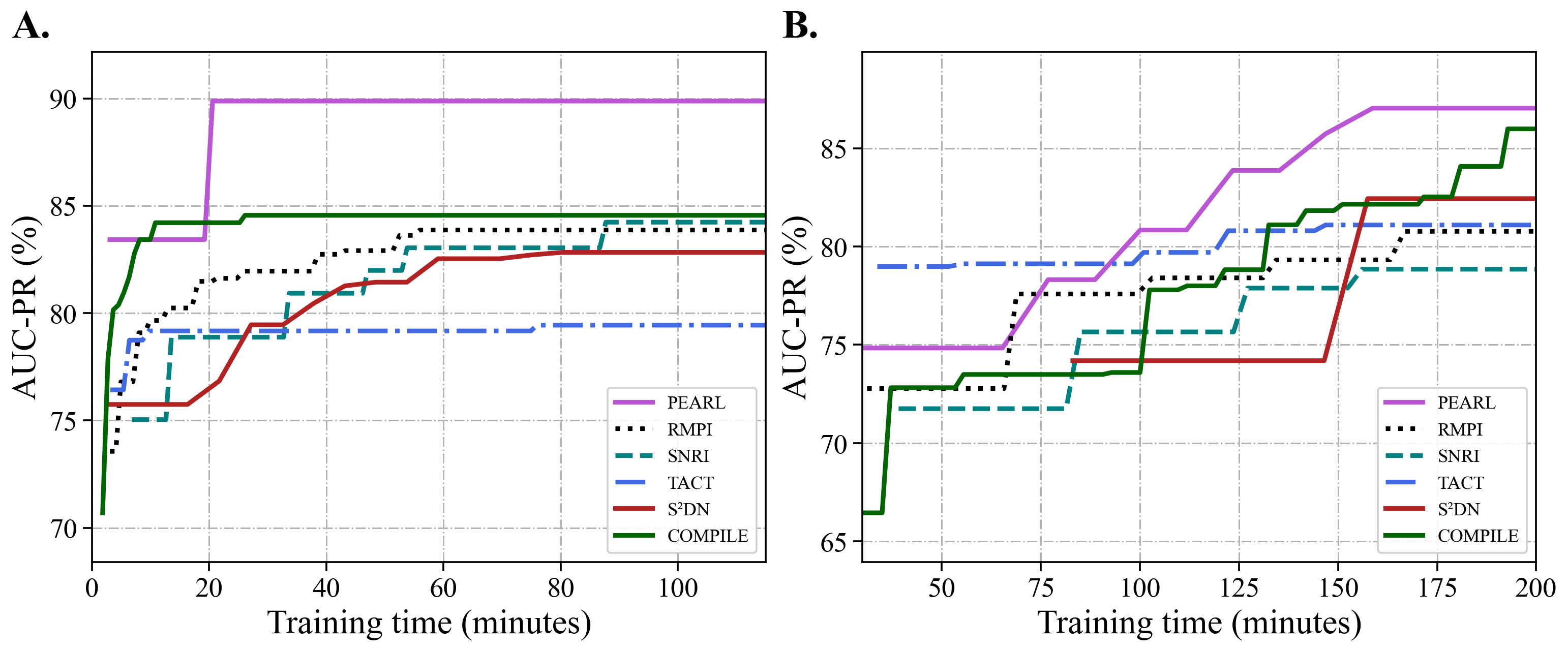}
    \caption{
Validation AUC-PR curves of PEARL and several competitive methods on FB15k-237 datasets.
(A) Training convergence curves on FB15k-237 V1;
(B) training convergence curves on FB15k-237 V3.
}
    \label{fig:auc_time}
\end{figure}

For training efficiency, we track the validation performance throughout the training process. Specifically, we record the validation area under the precision--recall curve (AUC-PR) with respect to the accumulated training time. Fig.~\ref{fig:auc_time} presents the AUC-PR curves on the FB15k-237 V1 and V3 datasets. On both datasets, PEARL exhibits rapid performance improvement during the early stages of training and consistently achieves higher validation AUC-PR than competing methods under comparable training time budgets. Moreover, PEARL converges faster than most baselines and ultimately attains the highest validation performance. These results indicate that
the proposed contextual subgraph modeling and path-aware reasoning
framework supports both accurate prediction and effective optimization. In addition, Table~\ref{tab:costs} reports the peak GPU memory consumption during training. PEARL requires the lowest GPU memory on FB15k-237 V1 and remains comparable to the most memory-efficient baseline on V3, demonstrating favorable memory efficiency. This efficiency is partly attributed to the LLM-guided path retrieval strategy, which reduces unnecessary path-level computations in subsequent reasoning.

For inference efficiency, Table~\ref{tab:costs} separately reports the preprocessing time and downstream inference time per query triple, together with the corresponding Hits@10 performance on FB15k-237 V1 and V3. The preprocessing time, reported as subgraph extraction time in the table, includes both contextual subgraph construction and LLM-guided path retrieval. The resulting contextual subgraph and retained relational paths are then directly used for subsequent neural inference. Although LLM-guided retrieval introduces additional preprocessing overhead, it removes redundant candidate paths and retains only informative reasoning chains, thereby reducing the complexity of downstream path modeling. As a result, PEARL maintains competitive downstream inference efficiency while achieving the best Hits@10 performance on both inductive settings.

\begin{table}[t]
    \centering
    \caption{The computational costs and prediction performance of several models on the FB15k-237 V1 and V3 datasets.}
    \label{tab:costs} 
    \renewcommand{\arraystretch}{1.2} 

    \begin{tabularx}{\linewidth}{lXXXXcccc} 
        \toprule
        \textbf{Model} 
        & \multicolumn{2}{c}{\textbf{\makecell[c]{Peak GPU \\ Memory \\(GB) }}}
        & \multicolumn{2}{c}{\textbf{\makecell[c]{Inference\\Time (ms)}}}
        & \multicolumn{2}{c}{\textbf{\makecell[c]{Subgraph \\ Extraction\\Time (ms)}}}
        & \multicolumn{2}{c}{\textbf{Hits@10}}\\
        \cmidrule(lr){2-3} \cmidrule(lr){4-5} \cmidrule(lr){6-7} \cmidrule(lr){8-9}
         & \textbf{V1} & \textbf{V3} & \textbf{V1} & \textbf{V3} & \textbf{V1} & \textbf{V3} & \textbf{V1} & \textbf{V3} \\
        \midrule
        CoMPILE 
        & \underline{1.33} & \textbf{5.57} & \textbf{23} & 104 & 90 & 120 & {64.73} & \underline{83.91} \\ 
        SNRI 
        & 2.50 & 13.47 & 80 & 192 & \underline{52} & \textbf{53} & {63.90} & {82.65} \\
        RMPI 
        & 3.83 & 15.33 & 64 & \textbf{70} & 111 & 144 & \underline{66.10} & {83.06} \\
        TACT 
        & 2.86 & 23.03 & 65 & \underline{72} & \textbf{50} & \underline{61} & {63.17} & {82.29} \\   
        S$^2$DN 
        & 2.03 & 9.49 & 250 & 570 & 59 & 84 & {64.39} & {82.94} \\
        \midrule
        PEARL 
        & \textbf{1.14} & \underline{5.59} & \underline{46} & 73 & 105 & 140 & \textbf{72.63} & \textbf{86.31} \\
        \bottomrule
    \end{tabularx}
\end{table}

\begin{figure*}[t]
    \centering
    \includegraphics[width=\linewidth]{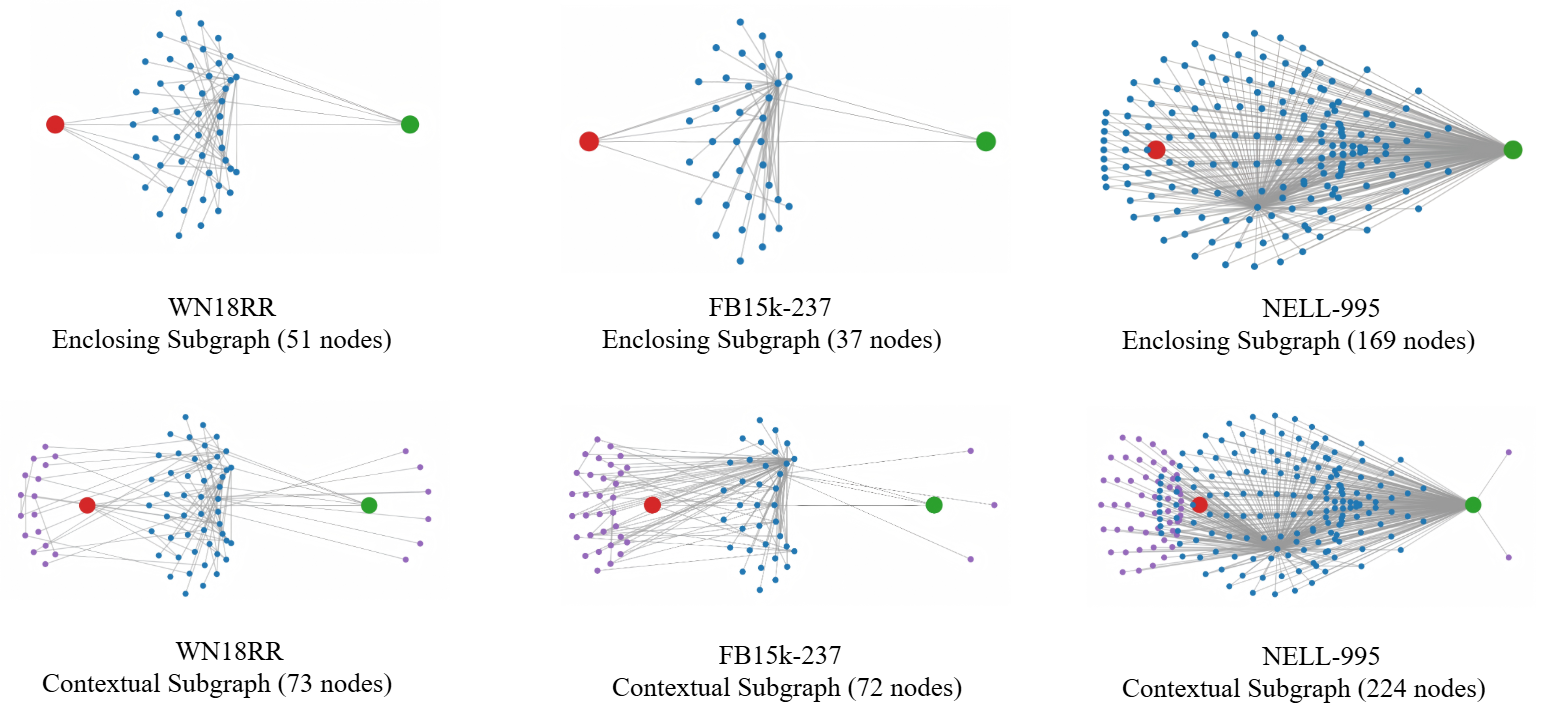}
    \caption{Comparison between enclosing subgraphs and contextual subgraphs on WN18RR, FB15k-237, and NELL-995, together with their corresponding numbers of nodes. The red and green nodes denote the head and tail entities of the query triple, respectively. Blue nodes represent the entities belonging to the intersection of the $k$-hop neighborhoods of the query entities, while purple nodes denote the additional contextual entities introduced from the neighborhood union.}
    \label{fig:comparsion_e_c}
\end{figure*}

\begin{figure*}[t]
\centering
\includegraphics[width=0.8\linewidth]{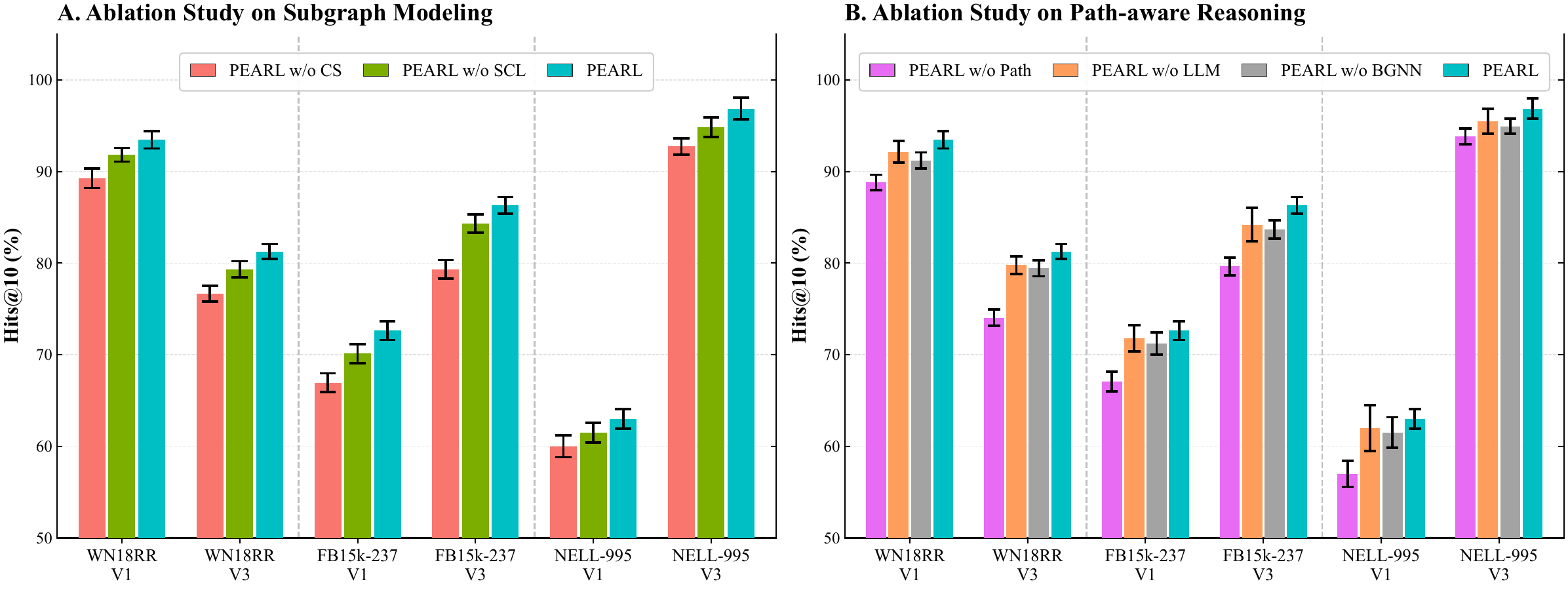}
\caption{
Ablation study of PEARL on WN18RR, FB15k-237, and NELL-995. Error bars represent scaled standard deviations. (A) Ablation on subgraph modeling, evaluating the effectiveness of contextual subgraph construction and subgraph contrastive learning. (B) Ablation on path-aware reasoning, including variants without relational path information, LLM-guided path retrieval, or the BGNN-based path–entity–global interaction mechanism.
}
\label{fig:ablation}
\end{figure*}

\subsection{Ablation Studies}

To comprehensively evaluate the contribution of each major design in PEARL, we conduct ablation studies from two complementary perspectives: (1) subgraph modeling and (2) path-aware reasoning. The corresponding results are summarized in Fig.~\ref{fig:ablation}.

\subsubsection{Ablation on Subgraph Modeling}

To evaluate the effectiveness of contextual subgraph modeling and subgraph contrastive learning, we compare the following variants:

\begin{itemize}

\item \textbf{PEARL w/o CS}: This variant replaces the contextual subgraph with the conventional enclosing subgraph and removes the subgraph contrastive learning objective.

\item \textbf{PEARL w/o SCL}: This variant retains the contextual subgraph but removes the subgraph contrastive learning objective.

\end{itemize}

As shown in Fig.~\ref{fig:ablation}(A), PEARL consistently outperforms both variants across all datasets. The performance gap between PEARL w/o CS and PEARL w/o SCL shows that introducing the contextual subgraph provides additional structural evidence beyond the conventional enclosing subgraph. Fig.~\ref{fig:comparsion_e_c} further illustrates their structural differences on WN18RR, FB15k-237, and NELL-995. Specifically, the contextual subgraph preserves the shared neighborhoods of the query entities while incorporating additional entities from the union of their $k$-hop neighborhoods, thereby providing broader structural context. Meanwhile, the further improvement of PEARL over PEARL w/o SCL demonstrates the benefit of subgraph contrastive learning in mitigating the noisy contextual information introduced by subgraph expansion. Additional visualization and quantitative analyses of the learned representations are provided in supplemental material D.

\begin{table*}[t]
\centering
\caption{Case study of LLM-guided relational path ranking.}
\label{tab:llm_case}
\small
\setlength{\tabcolsep}{5pt}
\renewcommand{\arraystretch}{1.12}
\begin{tabular}{p{0.24\textwidth} p{0.52\textwidth} cc}
\toprule
Query Triple & Candidate Relation Path & Importance Score & Rank \\
\midrule

\multirow{5}{=}{(Conan O'Brien, gender, Male)}
& Profession link $\rightarrow$ gender & 0.95 & 1 \\
& Marriage link $\rightarrow$ gender & 0.90 & 2 \\
& Friendship link $\rightarrow$ gender & 0.85 & 3 \\
& Religion link $\rightarrow$ film $\rightarrow$ gender & 0.80 & 4 \\
& Influenced by $\rightarrow$ gender & 0.75 & 5 \\

\midrule

\multirow{3}{=}{(Beasts of the Southern Wild, film release region, Japan)}
& film release region $\rightarrow$ location contains $\rightarrow$ film release region & 0.95 & 1 \\
& Netflix genre titles $\rightarrow$ film release region & 0.60 & 2 \\
& Film release region (Continent) $\rightarrow$ location adjoins $\rightarrow$ film release region & 0.55 & 3 \\

\bottomrule
\end{tabular}
\end{table*}

\begin{figure}[t]
    \centering
    \includegraphics[width=0.8\linewidth]{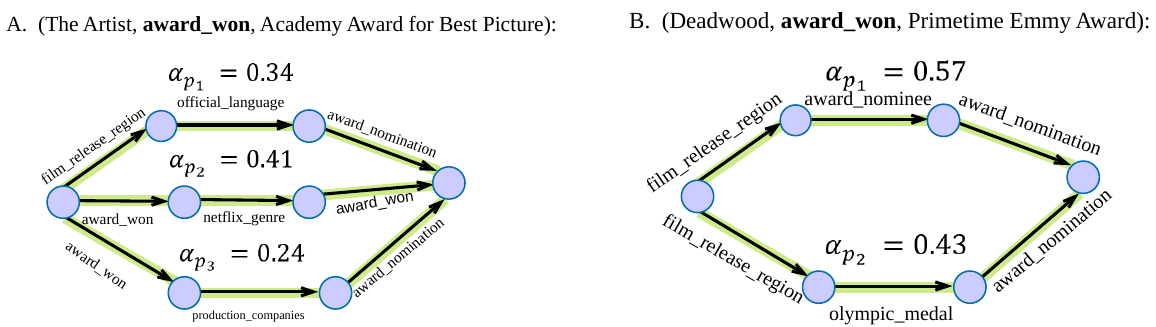}
    \caption{
Visualization of path attention weights for two query triples sharing the same target relation (\textit{award\_won}) on FB15k-237. 
}
    \label{fig:case}
\end{figure}

\subsubsection{Ablation on Path-aware Reasoning}

To further investigate the contribution of relational path reasoning, we compare the following variants:

\begin{itemize}

\item \textbf{PEARL w/o Path}: This variant removes relational path information and performs prediction using only the contextual subgraph representation.

\item \textbf{PEARL w/o LLM}: This variant removes the LLM-guided path retrieval module and randomly selects $M$ paths from the extracted candidate paths for subsequent reasoning.

\item \textbf{PEARL w/o BGNN}: This variant removes the BGNN module and directly uses the initial embeddings of the retrieved paths for prediction.

\end{itemize}

As shown in Fig.~\ref{fig:ablation}(B), removing relational path information consistently degrades performance, confirming that multi-hop paths provide complementary reasoning evidence beyond contextual subgraphs. Replacing LLM-guided path retrieval with random selection of $M$ candidate paths also leads to lower performance and larger standard deviations, indicating that semantic path retrieval helps identify more informative reasoning evidence and improves robustness compared with unguided path selection. Among the three variants, removing BGNN results in the largest performance degradation, highlighting the importance of explicitly modeling interactions between relational paths and contextual structural information. Overall, these results demonstrate that path information, LLM-guided retrieval, and path--entity interaction play complementary roles in the proposed path-aware reasoning framework.

\subsection{Case Study}

\subsubsection{LLM-guided Path Retrieval}

To assess the effectiveness of the proposed LLM-guided path retrieval module, we analyze representative query triples together with their candidate relational paths. The examples presented in Table~\ref{tab:llm_case} illustrate how the LLM estimates the semantic relevance of candidate paths by assigning importance scores with respect to the query triple.

For the query triple $(\textit{Conan O'Brien}, \textit{gender}, \textit{Male})$, the highest importance scores are assigned to paths involving \textit{Profession link}, \textit{Marriage link}, and \textit{Friendship link}, since these relations provide direct semantic evidence for inferring the  entity's gender. In contrast, paths related to \textit{Religion} and \textit{Influenced by} receive considerably lower scores due to their weaker semantic relevance.

A similar trend can be observed for the query triple $(\textit{Beasts of the Southern Wild}, \textit{film\ release\ region}, \textit{Japan})$. The path \textit{Film release region} $\rightarrow$ \textit{location contains} $\rightarrow$ \textit{film release region} obtains the highest importance score because it directly captures the geographical containment relation required for predicting the release region. By comparison, the remaining candidate paths provide only indirect or less relevant evidence, resulting in substantially lower importance scores.

These examples indicate that the LLM effectively distinguishes informative reasoning paths from semantically weak candidates. Consequently, the subsequent BGNN operates on a more reliable set of relational paths, reducing the influence of noisy reasoning signals while preserving the most discriminative logical evidence.

\subsubsection{Relation-Aware Path Attention Analysis}

To further investigate how structural context influences the interpretation
of relational paths, we analyze two query triples on FB15k-237.
They share the same target relation but involve different entity pairs:
(\textit{The Artist}, \textit{award\_won},
\textit{Academy Award for Best Picture}) and
(\textit{Deadwood}, \textit{award\_won},
\textit{Primetime Emmy Award}). Since both queries share the same  relation, the RAPF query vector remains identical, allowing us to isolate the influence of contextual subgraph structures on path importance.

As illustrated in Fig.~\ref{fig:case}, the two queries exhibit noticeably different attention distributions over candidate relational paths. For \textit{The Artist}, the path \textit{award\_won} $\rightarrow$ \textit{netflix\_genre} $\rightarrow$ \textit{award\_won} receives the highest attention weight, suggesting that genre-related evidence plays the dominant role in this structural context. In contrast, the two paths involving \textit{award\_nomination} receive relatively lower attention.

For \textit{Deadwood}, however, the attention distribution changes substantially. The paths containing \textit{award\_nomination} become the dominant reasoning evidence, indicating that nomination-related information is more informative under this local structural context than genre similarity.

These observations demonstrate that the semantic contribution of a relational path is jointly determined by its relation sequence and the surrounding contextual subgraph. Even for the same triple relation, PEARL adaptively emphasizes different reasoning paths according to the local structural environment, thereby enabling context-conditioned path interpretation rather than context-invariant path reasoning. Moreover, the consistent effectiveness of PEARL across relations with different occurrence frequencies further verifies its robustness in exploiting relational evidence under diverse knowledge graph structures. Additional results are provided in supplemental material F.

\section{Conclusion}

This paper investigates inductive knowledge graph completion from the perspective of context-aware relational reasoning and proposes PEARL, a framework that jointly models contextual subgraphs and relational paths for inductive reasoning over unseen entities. Extensive experiments on WN18RR, FB15k-237, and NELL-995 demonstrate that PEARL achieves strong overall performance against competitive baselines. This confirms that jointly exploiting structural context and relational paths provides an effective paradigm for inductive knowledge graph completion. In particular, the LLM-guided retrieval mechanism enables PEARL to identify relation-relevant paths from noisy local structures, while the subsequent path--entity interaction further adapts their semantics to the query-specific context. This design provides a unified framework for integrating LLM-derived semantic guidance with context-aware graph reasoning. Nevertheless, the current framework relies on offline LLM-based path retrieval and has not yet been evaluated on dynamic or temporal knowledge graphs. Future work will focus on more efficient adaptive path retrieval and extending PEARL to continual and temporal inductive reasoning scenarios.

\clearpage
\bibliographystyle{IEEEtran}
\bibliography{main}

\clearpage
\appendix
\section*{Supplemental Materials}
\begingroup
\renewcommand{\cite}[1]{\unskip}

\section{LLM-guided Path Retrieval Prompt}
\label{llm:retrieval}

To improve the quality of relational path reasoning, PEARL employs Qwen3-4B-Instruct as a semantic path retriever. Given a query triple and a set of candidate relational paths extracted from the contextual subgraph, the LLM is instructed to evaluate the semantic relevance of each candidate path with respect to the query triple by assigning an importance score. The candidate paths are then ranked according to these scores, and the resulting ranking is used to prioritize informative reasoning evidence. Table~\ref{tab:llm_prompt} presents the prompt template used for path scoring and ranking.

\begin{table}[H]
\centering
\caption{Prompt Template for LLM-guided Path Retrieval.}
\label{tab:llm_prompt}
\small
\begin{tabular}{p{1.8cm}|p{14cm}}
\toprule
\textbf{Role} & \textbf{Content} \\
\midrule

\textbf{System}
&




You are an expert in knowledge graphs, specializing in relation reasoning and path ranking.

Given a query triple and multiple candidate relational paths, assign an importance score in the range [0,1] to each candidate path according to how strongly it supports the query triple. A higher score indicates stronger semantic support.

Return a valid JSON object containing the candidate path indices together with their corresponding importance scores, sorted in descending order of score.

For example:

\begin{verbatim}
{
"path_ranking":[
{"index":0,"score":0.97},
{"index":3,"score":0.88},
{"index":1,"score":0.73}
]
}
\end{verbatim}

The returned indices must correspond exactly to the provided candidate paths. Do not generate non-existent indices or additional paths.
\\
\midrule

\textbf{User}
&
The input contains:
(1) the query head entity $h$,
(2) the query relation $r$,
(3) the query tail entity $t$, and
(4) a set of candidate relational paths represented as relation sequences from $h$ to $t$.
Evaluate the semantic relevance of each candidate path to the query triple by assigning an importance score, and rank all candidate paths from the strongest to the weakest supporting evidence.
Only use the provided candidate paths. Do not invent new paths. Only JSON output is allowed.
\\
\bottomrule
\end{tabular}
\end{table}

\section{Dataset and Evaluation Protocol}
We evaluate PEARL on three widely used IKGC benchmarks: WN18RR, FB15k-237, and NELL-995. Following GraIL~\cite{grail}, each dataset is divided into four inductive subsets (V1--V4) of increasing scale. For each subset, the training and test graphs are constructed under a shared global relation vocabulary while containing disjoint entity sets.
Table~\ref{tab:benchmarks} reports the detailed statistics of all datasets, including the numbers of relations, entities, and triples in the training and test graphs.

We follow the standard inductive evaluation setting introduced by GraIL~\cite{grail}. During training, validation, and testing, reverse relations are added following prior work. To ensure efficient model selection, we evaluate checkpoints on the validation set using the Area Under the Precision--Recall Curve (AUC-PR). Specifically, positive triples and sampled negative triples are used to construct a binary classification task, and the checkpoint with the highest validation AUC-PR is selected for final testing. For the final evaluation, we adopt the filtered ranking protocol and report Hits@1, Hits@10, and Mean Reciprocal Rank (MRR). All reported results are averaged over five independent runs, and the corresponding standard deviations are reported.
\begin{table}[t] \centering
\footnotesize
\caption{Inductive Benchmark Statistics. (REL: Relations, ENT: Entities, TR: Triples).}
\label{tab:benchmarks}
\resizebox{\columnwidth}{!}{ 
\begin{tabular}{ll rrr rrr rrr}
\toprule
\multirow{2}{*}{Version} & \multirow{2}{*}{Split} & \multicolumn{3}{c}{WN18RR} & \multicolumn{3}{c}{FB15k-237} & \multicolumn{3}{c}{NELL-995} \\
\cmidrule(lr){3-5} \cmidrule(lr){6-8} \cmidrule(lr){9-11}
& & REL & ENT & TR & REL & ENT & TR & REL & ENT & TR \\
\midrule
V1 & train & 9 & 2,746 & 6,678 & 183 & 2,000 & 5,226 & 14 & 10,915 & 5,540 \\
   & test  & 9 & 922   & 1,991 & 146 & 1,500 & 2,404 & 14 & 225    & 1,034 \\
\midrule
V2 & train & 10 & 6,954 & 18,968 & 203 & 3,000 & 12,085 & 88 & 2,564 & 10,109 \\
   & test  & 10 & 2,923 & 4,863  & 176 & 2,000 & 5,092  & 79 & 4,937 & 5,521 \\
\midrule
V3 & train & 11 & 12,078 & 32,150 & 218 & 4,000 & 22,394 & 142 & 4,647 & 20,117 \\
   & test  & 11 & 5,084  & 7,470  & 187 & 3,000 & 9,137  & 122 & 4,921 & 9,668 \\
\midrule
V4 & train & 9 & 3,861 & 9,842  & 222 & 5,000 & 33,916 & 77 & 2,092 & 9,289 \\
   & test  & 9 & 7,208 & 15,157 & 204 & 3,500 & 14,554 & 61 & 3,294 & 8,520 \\
\bottomrule
\end{tabular}
} 
\end{table} 

\section{Details of Baseline Methods}
\label{app:baselines}

\subsubsection{Rule-based Methods}

\begin{itemize}
\item \textbf{NeuralLP}~\cite{NeuralLP}: An end-to-end differentiable rule learning framework based on TensorLog that learns logical rules for multi-hop reasoning.

\item \textbf{DRUM}~\cite{drum}: An extension of differentiable rule learning that employs tensor products to learn logical rules of varying lengths while sharing parameters across path lengths.
\end{itemize}
\subsubsection{Subgraph-based Methods}

\begin{itemize}
\item \textbf{GraIL}~\cite{grail}: A pioneering IKGC model that performs reasoning over enclosing subgraphs.

\item \textbf{CoMPILE}~\cite{compile}: 
A subgraph-based model that strengthens the message-passing mechanism by explicitly modeling the directional information of relations within the local structure.

\item \textbf{TACT}~\cite{tact}: 
A topology-aware inductive link prediction method that models correlations between relations within local subgraphs.

\item \textbf{RMPI}~\cite{rmpi}: A relational message passing framework for inductive link prediction that reasons over subgraphs surrounding target entity pairs.

\item \textbf{LCILP}~\cite{lcilp}: A locality-aware subgraph method that employs Personalized PageRank (PPR)-based local clustering to extract informative neighborhoods around target links.

\item \textbf{S$^2$DN}~\cite{s2dn}: A dual-view subgraph neural network that balances core and auxiliary structural information.

\item \textbf{MINES}~\cite{mines}: A neighbor-enhanced subgraph reasoning framework with message intercommunication mechanisms.
\end{itemize}

\subsubsection{Subgraph \& Path-based Methods}

\begin{itemize}
\item \textbf{NBFNet}~\cite{nbfnet}: A neural Bellman-Ford framework that generalizes path-based reasoning to subgraph-level inference via differentiable path aggregation over the enclosing subgraph.

\item \textbf{SNRI}~\cite{snri}: A subgraph neighbor representation learning method that captures semantic information of unseen entities through neighboring relations.

\item \textbf{RPC-IR}~\cite{rpc}: A relational path contrastive learning framework that enhances path representations via contrastive objectives.

\item \textbf{CATS}~\cite{cats}\footnotemark: An LLM-based IKGC framework that performs reasoning based on latent type constraints, relational paths, and neighboring facts.
\end{itemize}

\footnotetext{
For a fair comparison, we implement CATS using the same Qwen3-4B-Instruct model as PEARL.
}

\section{Further Analysis of Contextual Subgraph Modeling}
\label{app:scl_analysis}


To further examine the impact of SCL, Fig.~\ref{fig:scl} visualizes the learned subgraph representations on FB15k-237. Without SCL, contextual subgraph embeddings show larger deviations from the core structural semantics captured by enclosing subgraphs. In contrast, PEARL with SCL achieves better alignment while maintaining richer contextual information, indicating that contrastive learning effectively reduces the influence of noisy contextual perturbations.

We additionally employ Positive Consistency $\mathcal{C}_{\mathrm{pos}}$ and Separation $\mathcal{S}$~\cite{scl_score1,scl_score2} to quantitatively evaluate representation quality. For a mini-batch of $B$ query triples, let $\mathbf{m}_{cs}^{(b)}$ and $\mathbf{m}_{enc}^{(b)}$ denote the contextual and enclosing subgraph embeddings of the $b$-th query, respectively. Positive Consistency is defined as:

\begin{equation}
\mathcal{C}_{\mathrm{pos}}
=
\frac{1}{B}
\sum_{b=1}^{B}
\cos
\left(
\mathbf{m}_{cs}^{(b)},
\mathbf{m}_{enc}^{(b)}
\right),
\end{equation}

whereas Separation $\mathcal{S}$ is computed as:

\begin{equation}
\mathcal{S}
=
\mathcal{C}_{\mathrm{pos}}
-
\frac{1}{B(B-1)}
\sum_{b\neq c}
\cos
\left(
\mathbf{m}_{cs}^{(b)},
\mathbf{m}_{cs}^{(c)}
\right).
\end{equation}

As shown in Fig.~\ref{fig:scl}, PEARL achieves higher $\mathcal{C}_{\mathrm{pos}}$ and $\mathcal{S}$ than PEARL w/o SCL, demonstrating that the proposed contrastive objective improves semantic consistency with core structural patterns while preserving discriminative contextual representations.

\begin{figure}[t]
    \centering

    \includegraphics[width=0.82\linewidth]{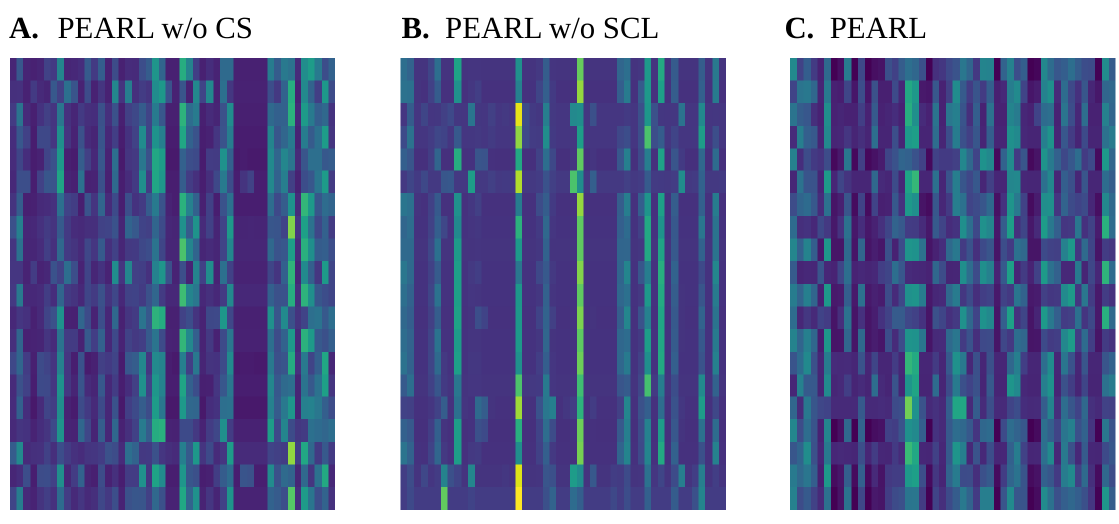}

    \vspace{0.8em}

    \small
    \renewcommand{\arraystretch}{1.1}
    \begin{tabular}{lcc}
    \toprule
    Variant & $\mathcal{C}_{\text{pos}}\uparrow$ & $\mathcal{S}\uparrow$ \\
    \midrule
    PEARL w/o SCL & 0.3950 & -0.3965 \\
    PEARL & \textbf{0.5424} & \textbf{-0.2507} \\
    \bottomrule
    \end{tabular}

   \caption{
Visualization and quantitative analysis of learned global subgraph embeddings on FB15k-237.
The upper part presents heatmaps of the subgraph embeddings learned by different variants:
(A) PEARL w/o Contextual Subgraph (PEARL w/o CS), which uses conventional enclosing subgraphs;
(B) PEARL w/o Subgraph Contrastive Learning (PEARL w/o SCL), which employs contextual subgraphs without contrastive regularization;
and (C) PEARL, which incorporates both contextual subgraph modeling and subgraph contrastive learning.
For a consistent comparison, all variants are evaluated on the same set of query triples.
The lower part reports the positive consistency ($C_{\mathrm{pos}}$) and separation ($S$) metrics.
}
\label{fig:scl}
\end{figure}

\section{Hyperparameter Sensitivity}
\label{hyperparameters}
In this subsection, we analyze the sensitivity of PEARL to different hyperparameter settings on the V1 subsets of WN18RR, FB15k-237, and NELL-995. The final hyperparameter configurations adopted for all datasets are summarized in Table~\ref{tab:hyperparameters}.

\textbf{Impact of batch size.}  
We first investigate the influence of batch size by varying it from $8$ to $64$. As shown in Fig.~\ref{fig:sen} (A), PEARL exhibits relatively stable performance across different batch sizes on all datasets. The best performance is achieved with a batch size of $8$ on WN18RR and NELL-995, while FB15k-237 obtains the optimal result with a batch size of $32$.

\textbf{Impact of learning rate.}  
We further evaluate the effect of the learning rate within the range from $0.0005$ to $0.01$. Fig.~\ref{fig:sen} (B) shows that excessively large learning rates slightly degrade the model performance on all datasets. PEARL achieves the best performance with a learning rate of $0.001$ on WN18RR and NELL-995, while FB15k-237 performs best with a smaller learning rate of $0.0005$. 

\textbf{Impact of subgraph size.}  
To investigate the influence of structural receptive fields, we vary the subgraph hop size $k$ from $1$ to $4$. As illustrated in Fig.~\ref{fig:sen} (C), PEARL achieves the best performance with $k=4$ on WN18RR, $k=3$ on FB15k-237, and $k=2$ on NELL-995. These results indicate that the optimal structural receptive field depends on the structural characteristics and density of different datasets.

\textbf{Impact of path number.}  
We study the influence of the number of relational paths by varying $M$ from $2$ to $5$. As shown in Fig.~\ref{fig:sen}(D), WN18RR and FB15k-237 achieve their best performance with $M=3$, while NELL-995 benefits from retaining more paths and performs best with $M=5$. Accordingly, we set $M=3$ for WN18RR and FB15k-237, and $M=5$ for NELL-995 in our experiments.

\textbf{Impact of loss balancing coefficients.}
We  investigate the influence of the loss balancing coefficients $\lambda_1$ and $\lambda_2$, which control the contributions of the primary link prediction loss $\mathcal{L}_{task}$ and the subgraph contrastive loss $\mathcal{L}_{SCL}$, respectively. Fig.~\ref{fig:v1lamda1} reports the Hits@10 performance under different coefficient combinations on WN18RR V1, FB15k-237 V1, and NELL-995 V1. For WN18RR, the best performance is achieved with a relatively small contrastive weight. In contrast, FB15k-237 and NELL-995
favor larger values of $\lambda_2$, indicating a stronger benefit from contrastive regularization on these datasets.

\section{Impact of Relation Frequency}
\label{app:relation_frequency}

To further investigate the effectiveness of PEARL under varying structural conditions, we conduct a fine-grained analysis based on the frequency of query relations in the training set. We hypothesize that high-frequency relations tend to exhibit more diverse and complex local subgraph patterns, where the benefit of contextual modeling becomes more pronounced.

Specifically, we partition test triples into three groups according to the training-set frequency of their query relations: bottom 25\%, middle 50\%, and top 25\%. We then evaluate PEARL and several competitive baselines on the V1 subsets of WN18RR, FB15k-237, and NELL-995. As reported in Table~\ref{table:freq}, PEARL achieves the best or competitive performance across most relation-frequency groups.

The gains are particularly evident in the high-frequency group of WN18RR and the medium- and high-frequency groups of FB15k-237. On NELL-995, PEARL achieves the best performance in the low- and high-frequency groups while remaining competitive in the middle-frequency group, indicating its effectiveness across different relation-frequency regimes.

Overall, these results demonstrate the robustness of PEARL across
different relation-frequency regimes and show that the proposed
framework can effectively exploit relational and structural evidence
under diverse frequency distributions. By anchoring relational paths within the contextual subgraph, PEARL adaptively leverages informative structural signals even when the distribution of relational patterns varies, leading to more reliable inductive reasoning.
\begin{table}[H]
\centering
\footnotesize  
\caption{DATASET-SPECIFIC HYPERPARAMETER SETTINGS OF PEARL.}
\label{tab:hyperparameters}
\begin{tabular}{l|ccc}   
\toprule
\textbf{Hyperparameters} & \textbf{WN18RR} & \textbf{FB15k-237} & \textbf{NELL-995} \\
\midrule
Subgraph hop size $k$ & 4     & 3         & 2        \\
learning rate   & 0.001  & 0.0005     & 0.001    \\
batch size      & 8      & 32        & 8        \\
Loss weights $(\lambda_1,\lambda_2)$ & 1.0, 0.2 & 0.6, 0.2 & 0.8, 0.6 \\
Number of retained paths $M$     & 3      & 3         & 5        \\
\bottomrule
\end{tabular}
\end{table}

\begin{figure*}[!t]
    \centering
    \includegraphics[
        width=\textwidth,        height=0.35\textheight,
        keepaspectratio
    ]{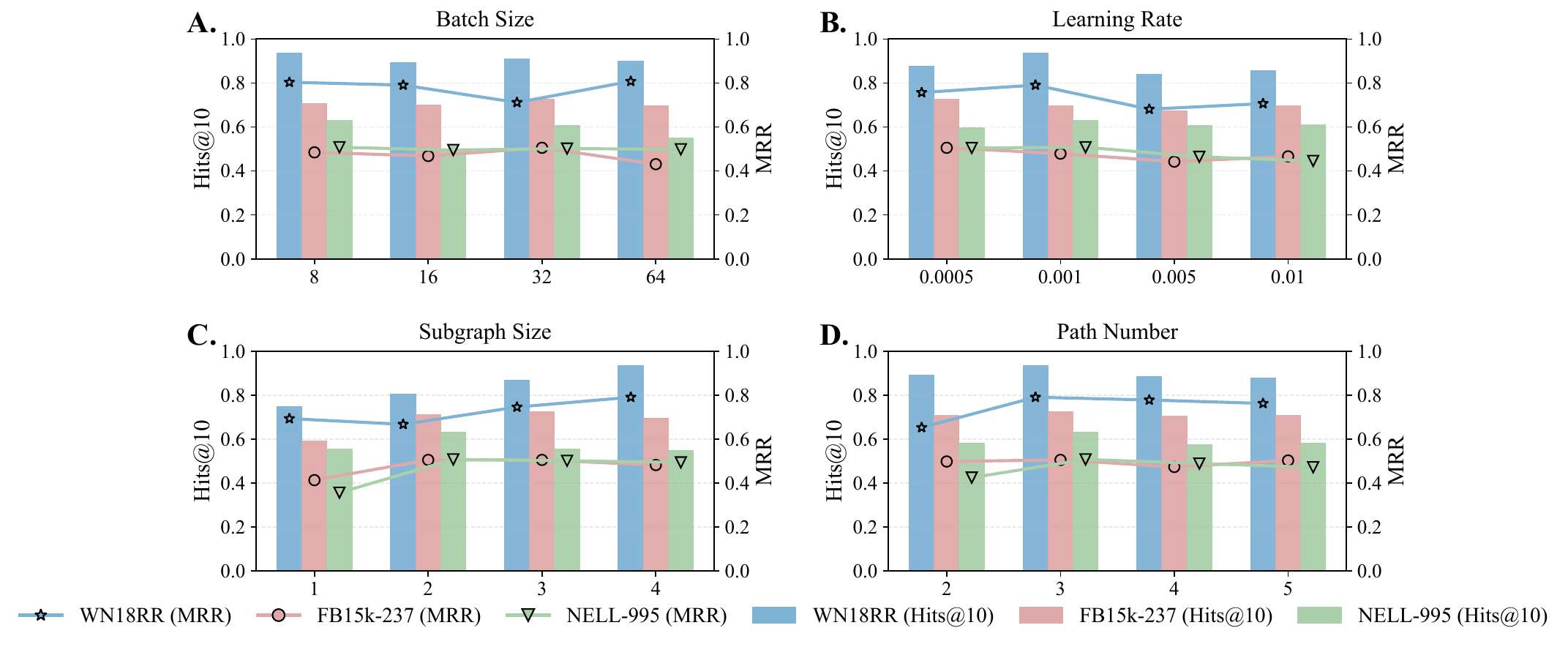}
    \caption{
        Hyperparameter sensitivity analysis of PEARL on WN18RR V1,
        FB15k-237 V1, and NELL-995 V1.
        (A) Effect of batch size;
        (B) effect of learning rate;
        (C) effect of subgraph size $k$;
        and (D) effect of the number of retrieved relational paths.
    }
    \label{fig:sen}
\end{figure*}

\begin{figure*}[!p]
    \centering
    \includegraphics[
        width=\textwidth,
        height=0.42\textheight,
        keepaspectratio
    ]{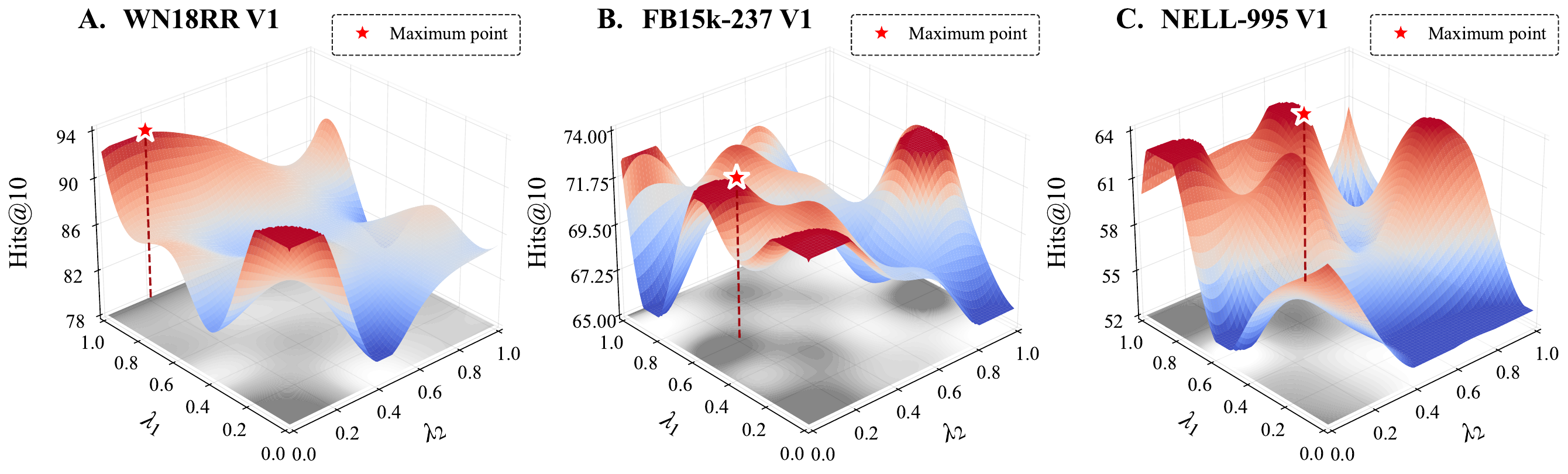}
    \caption{
        Hyperparameter sensitivity analysis with respect to the loss
        weights $\lambda_1$ and $\lambda_2$.
        (A) Hits@10 on WN18RR V1.
        (B) Hits@10 on FB15k-237 V1.
        (C) Hits@10 on NELL-995 V1.
    }
    \label{fig:v1lamda1}
\end{figure*}

\begin{table*}[!p]
    \centering
    \small
    \setlength{\tabcolsep}{4pt}
    \renewcommand{\arraystretch}{1.15}

    \caption{
        Impact of relation frequency on inductive link prediction
        (Hits@10, \%).
        Relations are grouped into the bottom 25\%, middle 50\%, and
        top 25\% according to their occurrence frequency in the training
        set. The best results are highlighted in bold.
    }
    \label{table:freq}

    \resizebox{\textwidth}{!}{%
\begin{tabular}{lccc|ccc|ccc}
        \toprule
        \multirow{2}{*}{\textbf{Methods}}
        & \multicolumn{3}{c|}{\textbf{WN18RR V1}}
        & \multicolumn{3}{c|}{\textbf{FB15k-237 V1}}
        & \multicolumn{3}{c}{\textbf{NELL-995 V1}} \\

        \cmidrule(lr){2-4}
        \cmidrule(lr){5-7}
        \cmidrule(lr){8-10}

        &
        \textbf{Bottom 25\%}
        & \textbf{Middle 50\%}
        & \textbf{Top 25\%}
        & \textbf{Bottom 25\%}
        & \textbf{Middle 50\%}
        & \textbf{Top 25\%}
        & \textbf{Bottom 25\%}
        & \textbf{Middle 50\%}
        & \textbf{Top 25\%} \\

        \midrule
        CoMPILE
        & 100.00 & 92.30 & 82.75
        & 52.61 & 53.84 & 72.01
        & 58.47 & 58.33 & 56.89 \\

        SNRI
        & 100.00 & \textbf{96.15} & 84.77
        & 39.47 & 46.15 & 71.64
        & 50.00 & 56.89 & 52.54 \\

        RMPI
        & 100.00 & 92.31 & 83.33
        & 39.47 & 39.42 & 60.45
        & 79.17 & 55.17 & 58.86 \\

        TACT
        & 100.00 & 92.31 & 83.33
        & 39.47 & 55.77 & 72.76
        & 54.17 & 56.90 & 55.08 \\

        S$^2$DN
        & 100.00 & 94.61 & 85.63
        & 49.47 & 55.26 & 71.26
        & 75.00 & \textbf{60.34} & 61.01 \\

        \midrule
        \textbf{PEARL}
        & \textbf{100.00}
        & \textbf{96.15}
        & \textbf{89.65}
        & \textbf{57.89}
        & \textbf{59.61}
        & \textbf{77.31}
        & \textbf{83.33}
        & 58.60
        & \textbf{62.71} \\

        \bottomrule
    \end{tabular}%
}
\end{table*}
\endgroup
\end{document}